\documentclass[fleqn,a4paper,oneside,10pt]{article}
\usepackage[british]{babel}

\usepackage{booktabs}
\usepackage{multirow}
\usepackage{subcaption}
\usepackage{longtable}
\usepackage{tabu}

\usepackage{mathtools}
\usepackage{amsmath}
\usepackage{amssymb}

\usepackage{graphicx}

\usepackage{pdflscape}
\usepackage{placeins}

\usepackage{authblk}

\usepackage[superscript,biblabel,nomove]{cite}
\usepackage{caption}
\DeclareCaptionLabelSeparator{upright}{ | }
\usepackage{geometry}
\usepackage{setspace}
\usepackage{hyperref}
\hypersetup{hidelinks}

\newlength{\tocsep} 
\let\oldbibliography\thebibliography
\renewcommand{\thebibliography}[1]{%
\addcontentsline{toc}{section}{\hspace*{-\tocsep}\refname}%
\oldbibliography{#1}%
\setlength\itemsep{0pt}%
}
\RequirePackage{jabbrv}

\begin{document}
\title{Assessing robustness of radiomic features by image perturbation}
\author[1,2,3]{Alex Zwanenburg}
\author[1,2,3]{Stefan Leger}
\author[1,4]{Linda Agolli}
\author[1,4]{Karoline Pilz}
\author[1,2,3,4,5]{Esther G.C. Troost}
\author[1,3,5]{Christian Richter}
\author[1,3,4]{Steffen L\"{o}ck}

\affil[1]{OncoRay – National Center for Radiation Research in Oncology, Faculty of Medicine and University Hospital Carl Gustav Carus, Technische Universität Dresden, Helmholtz-Zentrum Dresden - Rossendorf, Dresden, Germany}
\affil[2]{National Center for Tumor Diseases (NCT), Partner Site Dresden, Germany: German Cancer Research Center (DKFZ), Heidelberg, Germany; Faculty of Medicine and University Hospital Carl Gustav Carus, Technische Universität Dresden, Dresden, Germany, and; Helmholtz Association / Helmholtz-Zentrum Dresden - Rossendorf (HZDR), Dresden, Germany}
\affil[3]{German Cancer Consortium (DKTK), Partner Site Dresden, and German Cancer Research Center (DKFZ), Heidelberg, Germany}
\affil[4]{Department of Radiotherapy and Radiation Oncology, Faculty of Medicine and University Hospital Carl Gustav Carus, Technische Universität Dresden, Dresden, Germany}
\affil[5]{Helmholtz-Zentrum Dresden - Rossendorf, Institute of Radiooncology – OncoRay, Dresden, Germany}
\date{}
\maketitle

\begin{abstract}
Image features need to be robust against differences in positioning, acquisition and segmentation to ensure reproducibility. Radiomic models that only include robust features can be used to analyse new images, whereas models with non-robust features may fail to predict the outcome of interest accurately. Test-retest imaging is recommended to assess robustness, but may not be available for the phenotype of interest. We therefore investigated 18 methods to determine feature robustness based on image perturbations. Test-retest and perturbation robustness were compared for 4032 features that were computed from the gross tumour volume in two cohorts with computed tomography imaging: I) 31 non-small-cell lung cancer (NSCLC) patients; II): 19 head-and-neck squamous cell carcinoma (HNSCC) patients. Robustness was measured using the intraclass correlation coefficient (1,1) (ICC). Features with ICC$\geq0.90$ were considered robust. The NSCLC cohort contained more robust features for test-retest imaging than the HNSCC cohort (73.5\% vs. 34.0\%). A perturbation chain consisting of noise addition, affine translation, volume growth/shrinkage and supervoxel-based contour randomisation identified the fewest false positive robust features (NSCLC: 3.3\%; HNSCC: 10.0\%). Thus, this perturbation chain may be used to assess feature robustness.
\end{abstract}

\section{Introduction}
Radiomics is the high-throughput quantitative analysis of medical imaging to facilitate model-based treatment decisions \cite{Kumar2012,Lambin2012}. It relies on the computation of image biomarkers (features) within a region of interest (ROI). Features quantify different aspects of the ROI, such as mean intensity, volume and texture heterogeneity. Variations in patient positioning, image acquisition and segmentation affect each feature to varying degrees \cite{Mackin2015,Yip2016}. If radiomic models use features that are not robust against such influences, they will perform poorly when applied to new data \cite{Lambin2017}. Assessing feature robustness is thus recommended to improve generalisability of radiomic models.

Non-robust image features are commonly identified using test-retest imaging \cite{tixier2012reproducibility,Leijenaar2013,Balagurunathan2014,VanVelden2016,Desseroit2017}. In test-retest imaging, the same region of interest is imaged twice within a time interval of minutes to days. Consequently, these two images are similar, but not identical, which allows the identification of non-robust features. After identification, non-robust features are excluded from further analysis.

Although the identification of robust features is important, implementing test-retest imaging for every radiomic study has been difficult to achieve for several reasons. First, feature robustness is dependent on the phenotype of interest as well as the imaging modality. This means that information concerning feature robustness cannot be transferred between studies on different phenotypes\cite{VanTimmeren2016} and modalities\cite{Leijenaar2013}. Furthermore, feature values depend on multiple factors, including the voxel size and discretisation used\cite{Hatt2015,Shafiq-ul-Hassan2017,Mackin2017}. Thus, even if a previous study determined feature robustness for a particular phenotype and modality, the results may not be transferable due to the use of different computational settings. Second, test-retest imaging may be difficult to obtain generally, as it is not part of the clinical routine. Acquiring test-retest imaging would thus require additional resources in terms of personnel and imaging time, and, potentially, an increased patient radiation dose. An alternative would be to use the appropriate publicly available test-retest data set, but such data are likewise sparse.

It would therefore be convenient if feature robustness against perturbations could be assessed from single images. To do so, we can use methods more prevalent in the deep learning computer vision field. Here, networks are constructed to be invariant to various perturbations, e.g. noise, rotation and translation \cite{Arel2010}. To achieve invariance, images are distorted by applying such perturbations, and are subsequently used as input data to develop deep learning models. The same principle may apply to the handcrafted features that are considered in this work. We hypothesise that perturbations of single images may successfully identify the majority of features that are not robust in test-retest imaging. The aim is thus to identify perturbations that minimise the number of false positive robust features, using robustness in test-retest imaging as reference.

\section{Results}
Two test-retest data sets of computed tomography (CT) images were assessed, namely: I) a publicly available non-small cell lung cancer (NSCLC) cohort of 31 patients; and II) an in-house head and neck squamous cell carcinoma (HNSCC) cohort of 19 patients.

\begin{figure}[ht]
\centering
\includegraphics[width=1.0\textwidth]{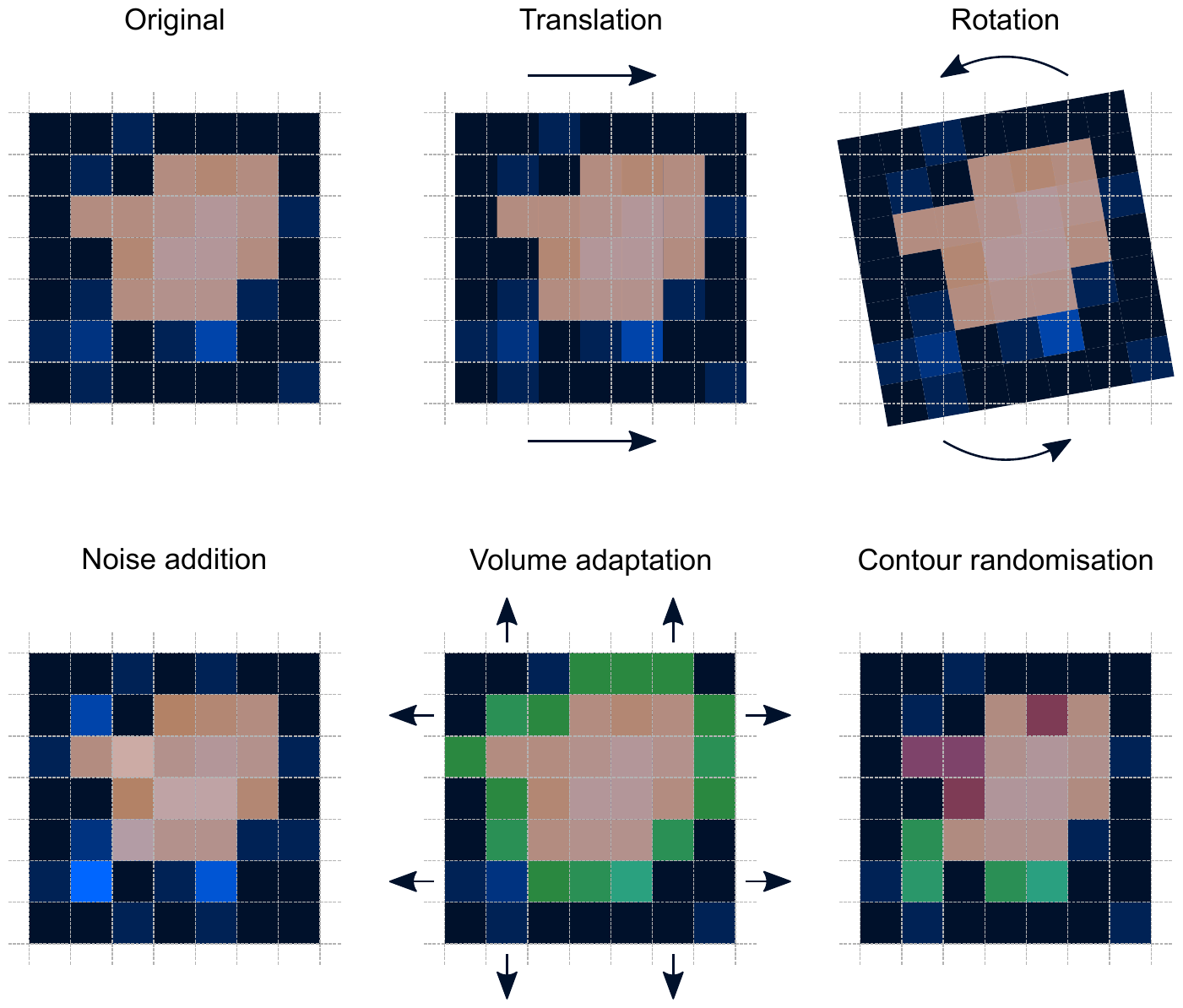}
\caption{Perturbation examples. To perturb an image (blue) and the region of interest mask (orange overlay), the original image is translated, rotated, noised, and has its mask adapted and randomised. Translation and rotation change both the image and its mask, whereas noise only distorts the image. Volume adaptation and contour randomisation change the mask by adding (green overlay) and removing voxels (red overlay). Note that translation and rotation require additional interpolation (not shown).}
\label{fig:perturbation}
\end{figure}

After delineating the gross tumour volume (GTV), the CT images were perturbed by rotation (R), Gaussian noise addition (N), translation (T), volume adaptation (growth/shrinkage of the ROI mask; V) and supervoxel-based contour randomisation (C), see Figure \ref{fig:perturbation} and Table \ref{table:perturbations}. Eighteen combinations of perturbations were created by chaining perturbation operations. All chains involved repetition with different settings or randomisation, and each instance generated a distorted image from which 4032 features were calculated.

\begin{figure}[ht]
\centering
\includegraphics[width=1.0\textwidth]{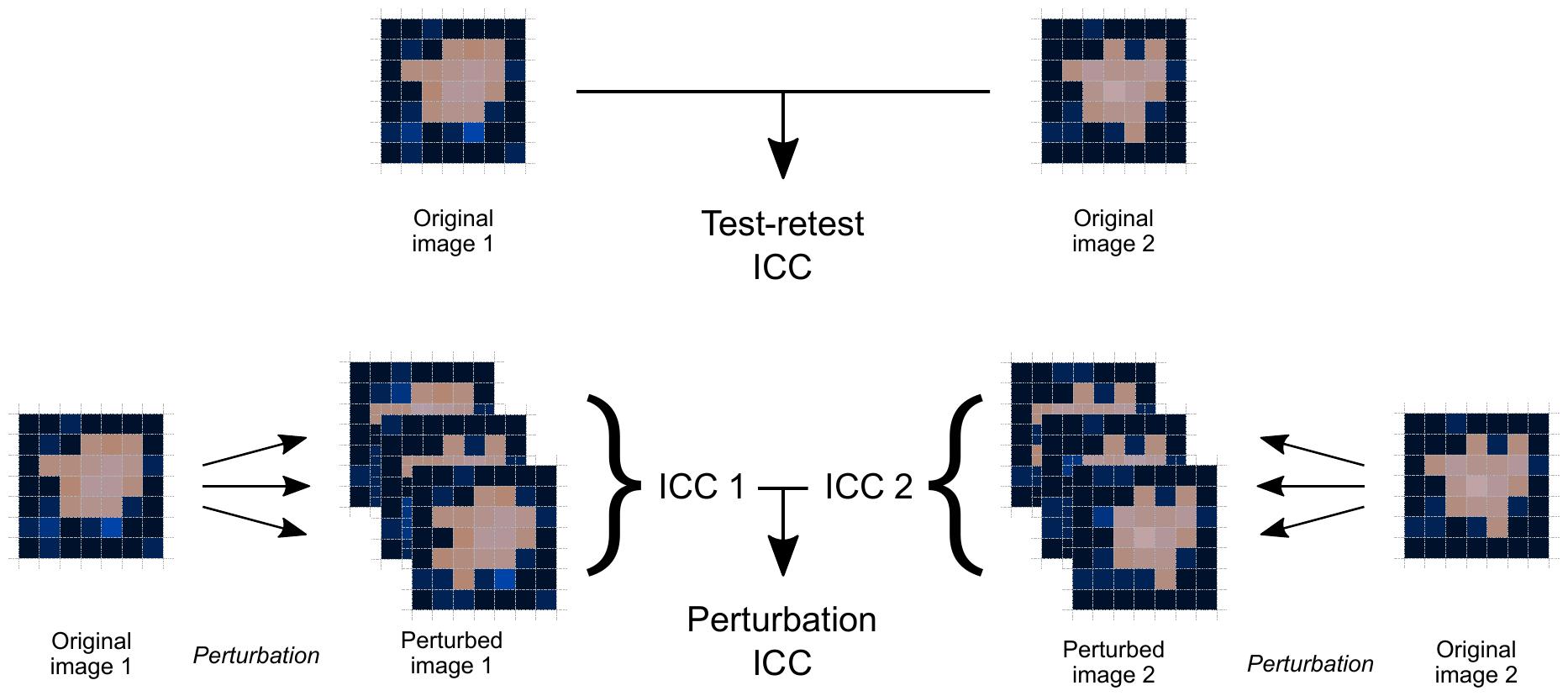}
\caption{Workflow to determine the test-retest and perturbation intraclass correlation coefficients (ICC) for each feature. The test-retest ICC was calculated directly between the same features in both images. To derive the perturbation ICC, an ICC was first calculated between feature values in perturbations of image 1 (ICC 1) and then again in perturbations of image 2 (ICC 2). The perturbation ICC is the average of ICC 1 and 2.}
\label{fig:workflow}
\end{figure}

Robustness of each feature was measured by the intraclass correlation coefficient (1,1) (ICC)\cite{Shrout1979}. We computed the ICC of a feature between either the test and retest images (test-retest ICC), or between the perturbed images for each perturbation chain (perturbation ICC), see Figure \ref{fig:workflow}. A feature was deemed robust if the estimated ICC exceeded the threshold value $\tau\geq 0.90$, and non-robust otherwise\cite{Bogowicz2016}. Test-retest and perturbation ICCs were calculated for all 4032 features.

% FIGURE Perturbation with examples
\newpage

% TABLE perturbations
\newpage
\begin{table}[ht]
\centering
\begin{tabu} to 0.99\textwidth {@{}X[4,l,p] X[1,c,p] X[1,c,p]@{}}
\toprule
\textbf{perturbation} & \textbf{abbreviation} & \textbf{\# perturbed images}\\
\midrule
rotation & R & 27\\
noise addition & N & 30\\
translation & T & 27\\
volume adaptation & V & 29\\
contour randomisation & C & 30\\
rotation and translation & RT & 32\\
rotation, noise addition and translation & RNT & 32\\
rotation and volume adaptation & RV & 30\\
rotation and contour randomisation & RC & 27\\
translation and volume adaptation & TV & 40\\
translation and contour randomisation & TC & 27\\
rotation, translation and contour randomisation & RTC & 32\\
rotation, noise addition, translation and contour randomisation & RNTC & 32\\
volume adaptation and contour randomisation & VC & 30\\
rotation, volume adaptation and contour randomisation & RVC & 30\\
rotation, noise addition, volume adaptation and contour randomisation & RNVC & 30\\
translation, volume adaptation and contour randomisation & TVC & 40\\
noise addition, translation, volume adaptation and contour randomisation & NTVC & 40\\
\bottomrule
\end{tabu}
\caption{List of perturbations, with their abbreviation and the number of different images generated by each perturbation. The settings used by each perturbation chain are listed in supplementary note 5}
\label{table:perturbations}
\end{table}

% FIGURE ICC calculation and comparison
\newpage

\subsection{Comparison between NSCLC and HNSCC cohorts}
To validate the basic premise that feature robustness is dependent on the phenotype, we compared feature robustness based on the test-retest ICC in both cohorts. In the NSCLC cohort 2963 (73.5\%) features were robust and 1069 (26.5\%) were non-robust. In the HNSCC cohort 1369 (34.0\%) features were robust and 2663 (66.0\%) non-robust. 1116 (27.7\%) and 816 (20.2\%) features were robust and non-robust in both cohorts, respectively. The robustness of the other 2100 (52.1\%) features was assessed differently between cohorts. 1847 (45.8\%) features were robust in the NSCLC cohort, but not in the HNSCC cohort, and 253 (6.3\%) features the other way around.

\subsection{Robustness under image perturbations}
\begin{figure}[ht]
\centering
\includegraphics[width=1.0\textwidth]{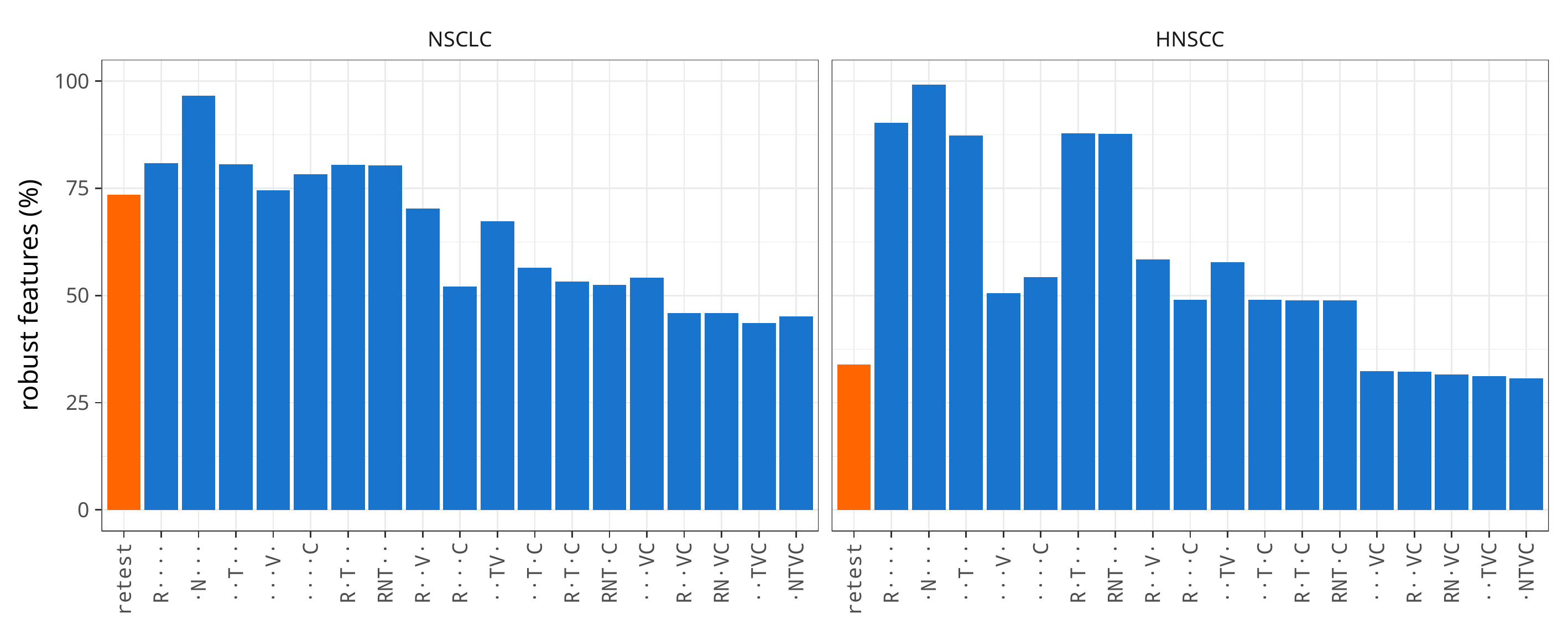}
\caption{Fraction of robust features identified for test-retest (orange) and perturbations (blue). Robustness was assessed using the intraclass correlation coefficient (ICC). Features with $\text{ICC}\geq 0.90$ were considered to be robust. Perturbations are abbreviated, see Table \ref{table:perturbations}: R: rotation; N: noise addition; T: translation; V: volume adaptation; C: contour randomisation.}
\label{fig:robust_features}
\end{figure}

The fraction of robust features for test-retest imaging and image perturbations is shown in Figure \ref{fig:robust_features}. In both cohorts, the N perturbation yielded the highest number of robust features (NSCLC: 96.6\%; HNSCC: 99.3\%), which was considerably higher than the number of robust features as determined by test-retest imaging (NSCLC: 73.5\%; HNSCC: 34.0\%). The lowest number of robust features in the NSCLC cohort was identified by the TVC perturbation chain (43.6\%), followed by NTVC (45.1\%), RNVC (45.9\%) and RVC (46.0\%). In the HNSCC cohort, NTVC (30.7\%), TVC (31.2\%), RNVC (31.6\%), RVC and VC (both 32.3\%) identified fewest robust features.

% FIGURE Number of robust features
\newpage

\subsection{Feature-wise comparison of perturbation and test-retest robustness}
\begin{figure}[ht]
\centering
\includegraphics[width=1.0\textwidth]{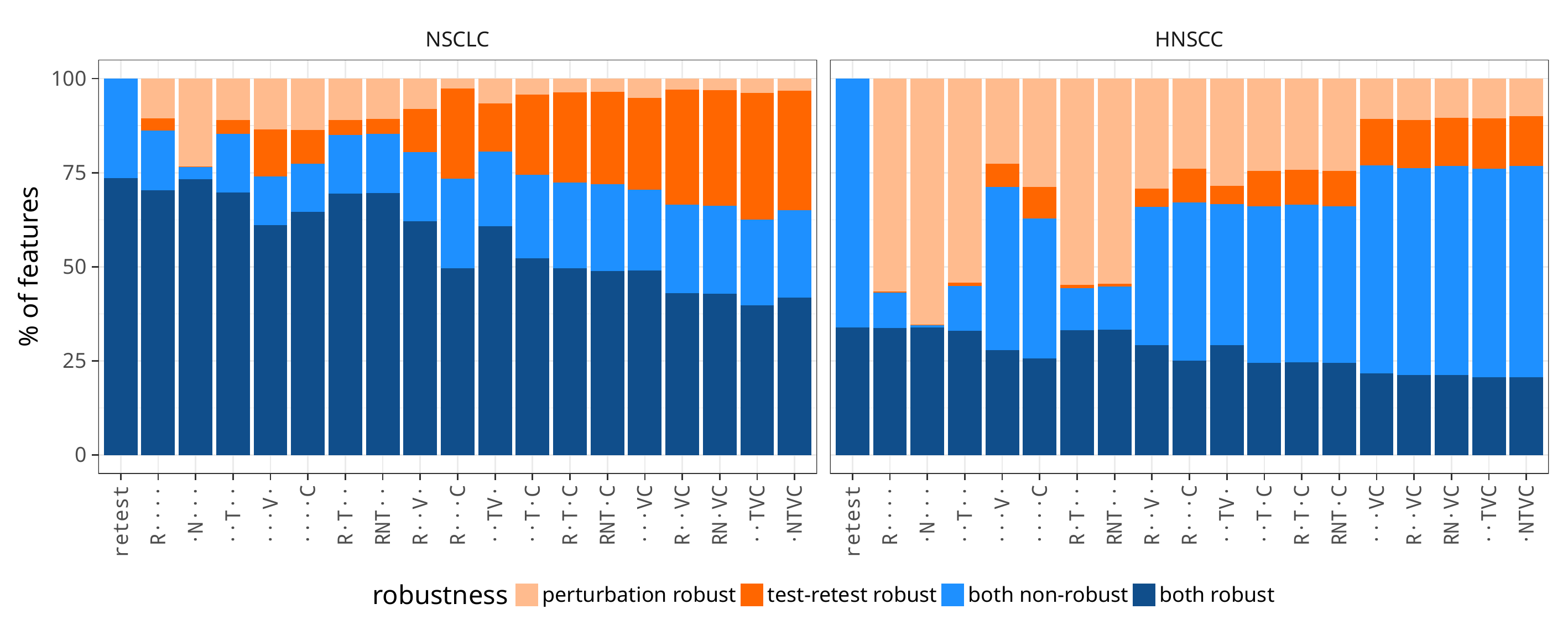}
\caption{Feature-wise comparison of robustness under test-retest and perturbation conditions. A feature was either robust under both test-retest and perturbation conditions (\textit{both robust}; true positive), non-robust under both conditions (\textit{both non-robust}; true negative), only robust under perturbations (\textit{perturbation robust}; false positive), or only robust under test-retest conditions (\textit{test-retest robust}; false negative). Test-retest robustness was used as reference, and the corresponding column only contains true positives and negatives. Robustness was assessed using the intraclass correlation coefficient (ICC). Features with $\text{ICC}\geq 0.90$ were considered to be robust. Perturbations are abbreviated, see Table \ref{table:perturbations}: R: rotation; N: noise addition; T: translation; V: volume adaptation; C: contour randomisation.}
\label{fig:robust_overlap}
\end{figure}

Test-retest and perturbation robustness were also compared directly for the same feature. A feature is either robust under both perturbation and test-retest conditions, non-robust under both, or robust under test-retest or perturbation conditions only. Using test-retest robustness as a reference, these conditions represent true positive, true negative, false negative and false positive cases, respectively. The direct comparison of robustness is presented in Figure \ref{fig:robust_overlap}.

No perturbation identified every feature that was non-robust under test-retest conditions. The number of false positives differed between perturbations and cohorts. Perturbation chains in the NSCLC cohort yielded less false positives than the HNSCC cohort on average (7.8\% vs. 30.3\%).

In the NSCLC cohort, the RC perturbation chain caused the lowest number of false positives (2.6\%), followed by RVC (3.0\%), RNVC (3.1\%) and NTVC (3.3\%). The lowest false positive fraction in the HNSCC cohort was produced by NTVC perturbation chain (10.0\%), followed by RNVC (10.4\%), TVC (10.6\%), VC (10.7\%) and RVC (11.1\%). In the HNSCC cohort, the RC perturbation chain led to 24.0\% false positives.

% FIGURE Overlap between robustness in test-retest and perturbation sets
\newpage

\section{Discussion}
We compared several methods for perturbing images to determine feature robustness. The chained perturbation that consists of noise addition, translation, volume adaptation and contour randomisation (NTVC) led to a low number of false positives in both cohorts, using test-retest robustness as reference. The TVC, RNVC and RVC perturbation chains showed similar performance. Hence any of these perturbation chains may be used to assess feature robustness.

Other perturbation methods performed poorly, particularly if only one kind of perturbation was used. This includes methods such as noise addition or simple rotations or translations. The combination of rotation and translation was not better than rotation or translation alone. Chaining methods that primarily alter the intensity content (noise, translation, rotation) with methods that update the region of interest mask (volume adaptation and contour randomisation) did improve results in terms of less false positives with regard to test-retest imaging.

We used test-retest imaging as a reference standard. However, test-retest imaging has its limitations. In particular, the number of test-retest images is usually just two, which may not suffice to determine the ICC with good precision \cite{Koo2016}. This uncertainty is reflected in the 95\% confidence interval of each ICC value. The average width of the 95\% confidence interval of test-retest ICCs was 0.12 (NSCLC) and 0.35 (HNSCC). Image perturbations can be repeated multiple times and thus allows a more precise estimation of the ICC. For instance, the average confidence interval width of the NTCV perturbation chain in the NSCLC cohort was similar to that of test-retest imaging, with a width of 0.11 for both CT\textsubscript{1} and CT\textsubscript{2}) images. However, the perturbation ICC in the HNSCC cohort could be determined considerably more precise with average confidence interval widths of 0.18 (CT\textsubscript{1}) and 0.17 (CT\textsubscript{2}). The large uncertainty in test-retest robustness for the HNSCC cohort may have contributed to a higher number of false positives.

Another limitation of using test-retest imaging as a reference is that the test-retest images may still be too similar. The same equipment and protocols may still be used, and segmentation may still be performed by a single expert. Thus it cannot be ruled out that some of the false negative features were correctly assessed as not robust by perturbation.

The above limitation may also explain the lower number of false negatives in the HNSCC cohort compared to the NSCLC cohort. Two different image acquisition protocols were used in the HNSCC cohort, whereas only one protocol was used for test-retest imaging in the NSCLC cohort. This is noticeable in the differences in exposure. The exposure between both HNSCC images differed by a factor 4 on average, whereas exposure in the NSCLC set was similar between images. The HNSCC test-retest set may thus have captured differences in exposure. However, the effect of exposure and tube current on feature robustness has been contested. Larue \textit{et al.} and Mackin \textit{et al.} both found that exposure had a marginal effect on feature robustness \cite{Larue2017,Mackin2018}, whereas Midya \textit{et al.} found that it had a more pronounced effect \cite{Midya2018}. Test-retest robustness may also have been affected by the difference in reconstruction kernels. Though both kernels in the HNSCC cohort produce smooth images, different reconstruction kernels may strongly affect feature values \cite{Zhao2016,He2016}.

The current study has some limitations. One limitation is that we only assessed test-retest imaging based on computed tomography, as test-retest data sets for other modalities were not available to us. The proposed methodology should be assessed for different modalities, e.g. positron emission tomography (PET) and magnetic resonance imaging (MRI).

Another limitation is that we did not assess delineation uncertainties. Delineation uncertainties also cause variability in feature values\cite{Pavic2018}. Volume adaptation and contour randomisation perturbations try to induce this uncertainty, but a comparison against a multiple delineation data set should be performed in the future.

Perturbations allow us to perform repeated measurements and it is important to reckon how this may be used for radiomic modelling. We consider three methods for incorporating repeated measurements into radiomic modelling. The first, straightforward, method is to include only robust features in the modelling process. This method is currently used when robustness is determined using test-retest imaging and its implementation into modelling workflows should therefore be easy. Moreover, this method is useful when only a subset of the development cohort is perturbed, or a separate data set is used for robustness analysis.

The second way to use repeated measurements for radiomic modelling is by averaging the measurements of each feature. Averaging effectively increases feature robustness as the corresponding (panel/multiple rater) ICC is always higher than that of a single measurement \cite{Shrout1979}. The mean values of the features that are robust according to the panel ICC are then included into the modelling process. This method requires that all images in the development cohort are perturbed, and is thus more computationally expensive than the first.

The final method builds upon the second, and is conceptually close to the use of image perturbations for deep learning. Instead of averaging values and selecting robust features prior to modelling, all values are included in the model development process. One advantage of this method is that information concerning the distribution of feature values within and across samples is not lost, and may be exploited during the model development process. Another advantage is that a robustness threshold is not required. However, this method does require that all images in the development cohort are perturbed and may add complexity to radiomic modelling frameworks. A future study should compare these three methods and its effect on the performance of radiomic models.

In conclusion, we investigated the use of image perturbations to determine the robustness of radiomic features, using test-retest imaging as reference. Our findings indicate that chained perturbations which perturb image intensity and segmentation may be used instead of test-retest imaging to determine feature robustness.

\section{Methods}

\subsection{Test-retest cohorts}
Two patient cohorts with test-retest computed tomography imaging were used: a publicly available non-small cell lung cancer cohort of 31 patients \cite{Zhao2009,Zhao2015} and an in-house cohort (DRKS 00006007) of 19 patients with locally advanced head and neck squamous cell carcinoma \cite{Loeck2017}. The NSCLC cohort is available from the Cancer Imaging Archive \cite{Clark2013}. For the NSCLC cohort, two separate images were acquired within 15 minutes of each other, using the same scanner and acquisition protocol. Images in the HNSCC cohort were acquired within 4 days of each other using a different protocol, i.e. one CT image was acquired for \textsuperscript{18}F-Fludeoxyglucose positron emission tomography (PET) attenuation correction, and the other for attenuation correction of \textsuperscript{18}F-fluoromisonidazole PET. Approval for analysis of the in-house data set was provided by the local ethics committee (EK 177042017). Image acquisition parameters for both cohorts are shown in supplementary note 1.

The GTV was delineated by experienced radio-oncologists (L.A., K.P., E.G.C.T) using the Raystation 4.6 treatment planning system software (RaySearch Laboratories AB, Stockholm, Sweden), and subsequently used as the region of interest.

\subsection{Image processing}
\begin{figure}[ht]
\centering
\includegraphics[scale=0.85]{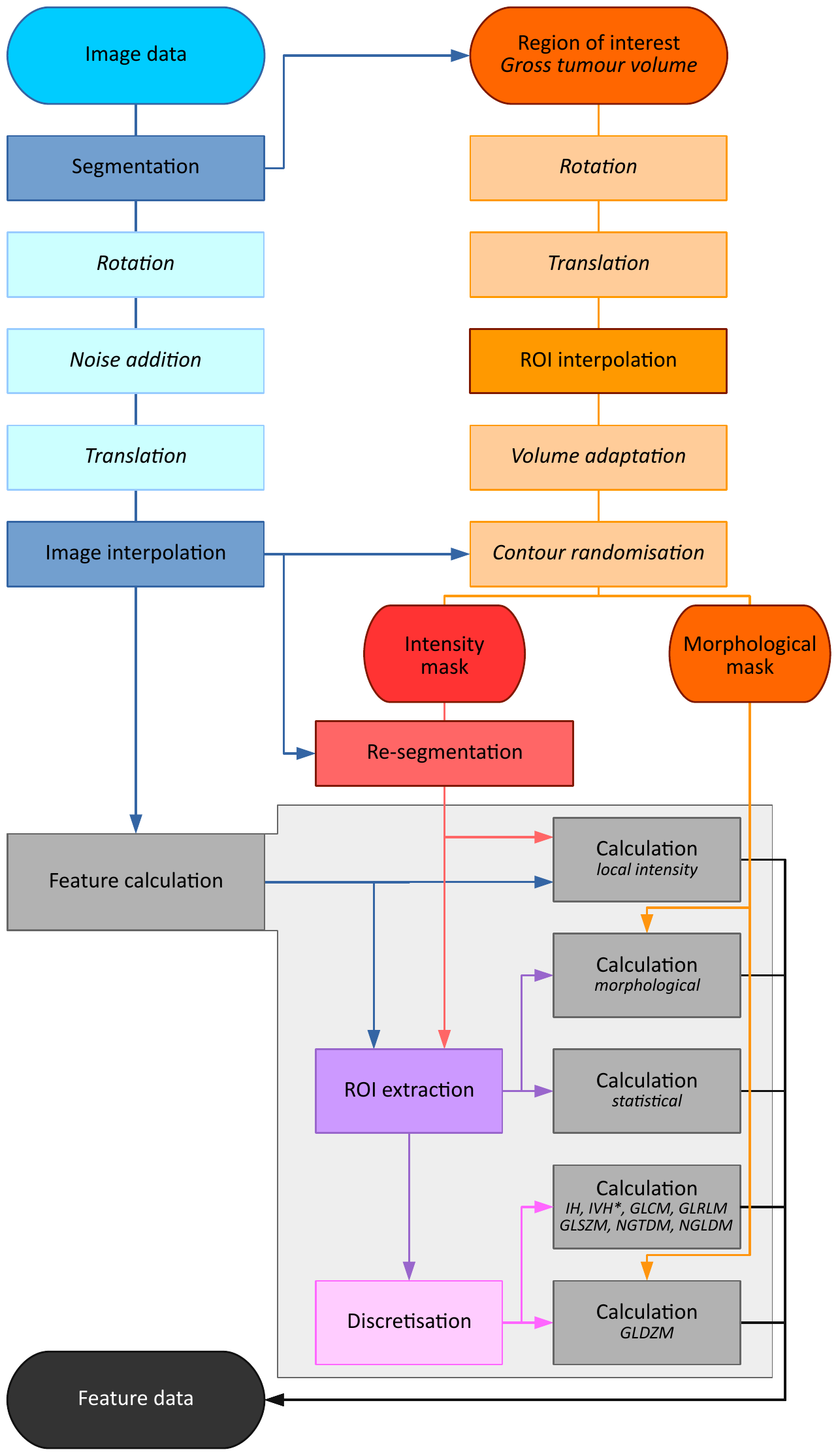}
\caption{Image processing scheme with perturbations. A computed tomography (CT) image and a segmented gross tumour volume (GTV) are used as the input image data and the region of interest (ROI) respectively. The CT and ROI are processed to compute image features. Rotation, translation, noise addition, volume adaptation and contour randomisation are optional perturbation steps. Other image processing steps are detailed in the documentation of the image biomarker standardisation initiative (IBSI)\cite{Zwanenburg2016}. IH: intensity histogram; IVH: intensity-volume histogram; GLCM: grey level co-occurrence matrix; GLRLM: grey level run length matrix; GLSZM: grey level size zone matrix; GLDZM: grey level distance zone matrix; NGDTM: neighbourhood grey tone difference matrix; NGLDM: neighbouring grey level dependence matrix. This figure is based on the image processing scheme in the IBSI document.}
\label{fig:image_processing}
\end{figure}

Image processing was conducted using the scheme and recommendations provided by the Image Biomarker Standardisation Initiative (IBSI)\cite{Zwanenburg2016}. An overview of the processing steps is provided in Figure \ref{fig:image_processing}, and further details may be found in the IBSI documentation. A complete overview of the image processing parameters, excluding perturbation-related parameters, may be found in Table \ref{table:image_processing_parameters}.

In short, after loading a CT image, \texttt{DICOM RTSTRUCT} polygons were used to generate a voxel-based segmentation mask for the GTV ROI. The image and mask were then both rotated over a set angle $\theta$ (optional). Gaussian noise, based on the noise levels present in the original image, was added to the image (optional). Subsequently, both image and mask were translated with a sub-voxel shift $\eta$ (optional) and interpolated with prior Gaussian anti-aliasing (supplementary note 2). After interpolation to isotropic voxel dimensions, the image intensity values were rounded to the nearest integer Hounsfield unit, and the mask was re-labeled based on the partial voxel volume threshold. The mask was then grown or shrunk to alter the volume by a fraction $\tau$ (optional), before being perturbed by supervoxel-based contour randomisation \cite{Achanta2012} (optional). The mask was subsequently copied to generate an intensity mask and a morphological mask. The intensity mask was re-segmented to an intensity range which includes only soft-tissue voxels. Voxels with intensities deviating more than three standard deviations from the mean of the ROI were excluded from the intensity mask as well \cite{Collewet2004,Vallieres2015}. The image and both masks were subsequently used to compute radiomic features, with several feature families requiring additional discretisation (supplementary note 3).

% FIGURE Image processing scheme
\newpage

% TABLE image processing parameters
\newpage
\begin{table}[ht]
\centering
\begin{tabu} to 0.99\textwidth {@{}X[3,l,p] X[1,c,p] X[1,c,p]@{}}
\toprule
\textbf{parameter} & \textbf{NSCLC} & \textbf{HNSCC}\\
\midrule
interpolated isotropic voxel spacing (mm) & $1, 2, 3, 4$ & $1, 2, 3, 4$ \\
pre-interpolation filter & gaussian, $\beta=0.93$ & gaussian, $\beta=0.93$\\
image interpolation method & trilinear & trilinear \\
image intensity rounding & to nearest HU & to nearest HU \\
ROI interpolation method & trilinear & trilinear \\
ROI mask partial volume threshold & 0.5 & 0.5\\
re-segmentation range (HU) & $\left[-300,200\right]$ & $\left[-150,180\right]$\\
re-segmentation outlier threshold & $\pm 3 \sigma$ & $\pm 3 \sigma$\\
discretisation  & &\\
\hspace{10mm} fixed bin number (bins) & 8, 16, 32, 64 & 8, 16, 32, 64\\
\hspace{10mm} fixed bin size (HU) & 6, 12, 18, 24 & 6, 12, 18, 24\\
\bottomrule
\end{tabu}
\caption{Image processing parameters for both NSCLC and HNSCC data sets. The isotropic voxel spacing is defined in three dimensions, i.e. a spacing of $2$ mm corresponds to a voxel dimension of $2\times2\times2$ mm. Discretisation was performed using two methods (\textit{fixed bin number} and \textit{fixed bin size}) with varying bin sizes. \textit{ROI}: region of interest; \textit{HU}: Hounsfield unit; $\sigma$: standard deviation of voxel intensities within the region of interest.}
\label{table:image_processing_parameters}
\end{table}

\subsection{Image perturbations}
Five basic image perturbation methods were implemented in the image processing scheme described above. These were rotation (R), noise addition (N), translation (T), volume adaptation (V) and contour randomisation (C). Examples are shown in Figure \ref{fig:perturbation}. Rotation perturbs the image and mask by performing an affine transformation that rotates the image and mask in the axial ($x$,$y$) plane, i.e. around the $z$-axis, for a specified angle $\theta \in [-13^{\circ},13^{\circ}]$. Noise addition perturbs image intensities by adding random noise that was drawn from a normal distribution with mean $0$ and a standard deviation equal to the estimated standard deviation of the noise present in the image. Translation perturbs the image and mask by performing an affine transformation that shifts the image and mask for specified fractions ($\eta \in [0.25,0.75]$) of the isotropic voxel spacing along the $x$, $y$ and $z$ axis. Volume adaptation grows and/or shrinks the mask by a specified fraction $\tau \in [-0.28,0.28]$. Contour randomisation is based on simple linear iterative clustering\cite{Achanta2012}, and perturbs the mask by randomly selecting supervoxels based on the overlap with the original mask. The algorithmic implementation of these perturbations is described in supplementary note 4.

Perturbations were chained using the settings documented in supplementary note 5. Each rotation angle and volume adaptation fraction led to generation of a new image. Noise addition and contour randomisation could be repeated multiple times, with each repetition producing a new perturbed image. The translation fraction was permuted over the different directions. For example, for translation fractions $\eta=\left\lbrace0.25, 0.5\right\rbrace$, $2^3=8$ permutations were generated. When chaining perturbations, all provided parameters were permuted.

An overview of the perturbation chains and the number of perturbed images generated is shown in Table \ref{table:perturbations}. All perturbation chains produced between 27 and 40 perturbed images.

\subsection{Features}
All features defined in the IBSI documentation were implemented \cite{Zwanenburg2016}, leading to a basic set of 182 features. These features were calculated at multiple scales, namely for isotropic voxel spacings of 1, 2, 3 and 4 mm\cite{Vallieres2017}. 118 features of the basic set required discretisation. Both fixed bin size and fixed bin width discretisation algorithms were used, each with four settings. Thus, 4032 features were computed in each image. Supplementary note 3 contains further details with regard to feature computation.

Both image processing and feature computation were conducted using our IBSI-compliant in-house framework based on Python 3.6\cite{Leger2017}.
% BENCHMARK LIST

\subsection{Robustness analysis}
Feature robustness was assessed using the intraclass correlation coefficient (1,1) (ICC)\cite{Shrout1979}, based on the assumption that test-retest images, as well as perturbations, possess no consistent bias. The highest possible ICC value is 1.00, which indicates that feature values are fully repeatable between test-retest images or perturbations. Lower values denote an increasing measurement variance with respect to the intra-patient variance, and thus lower repeatability. Image features with $\text{ICC}\geq 0.90$ were considered to be robust\cite{Bogowicz2016}, and non-robust otherwise.

The test-retest ICC was determined between both CT images, see Figure \ref{fig:workflow}. Perturbation ICCs were first computed separately for the test and retest images. Subsequently, perturbation ICCs were averaged over test and retest images to facilitate comparison with the test-retest ICC, as there was no consistent bias toward higher ICC values for one image set (see supplementary note 6).

Feature robustness was assessed using R 3.4.2 \cite{RCoreTeam2017} and ICCs were computed using code adapted from the \texttt{psych} R-package\cite{psych}.

\FloatBarrier

\newpage
\renewcommand{\thefigure}{S\arabic{figure}}
\setcounter{figure}{0}
\renewcommand{\thetable}{S\arabic{table}}
\setcounter{table}{0}
\section*{Supplementary note 1: image acquisition parameters}
Computed tomography (CT) images were acquired for both the non-small-cell lung carcinoma (NSCLC) and head and neck squamous cell carcinoma (HNSCC) cohorts. For the NSCLC cohort, a second CT image was acquired 15 minutes after the first acquisition. The patient was asked to leave the table between the scans and was repositioned before the second image acquisition. For the HNSCC cohort a second CT image was recorded to determine attenuation corrections for positron emission tomography (PET). This PET-CT scan was recorded within 4 days after the original diagnostic CT scan. Acquisition parameters and characteristics are shown in Table \ref{table:image_acquisition_parameters}.

\begin{table}[ht]
\centering
\footnotesize
\begin{tabu} to 0.99\textwidth {@{}X[1.8,l,p] X[1,c,p] X[1,c,p] X[1,c,p] X[1,c,p]@{}}
\toprule
\textbf{parameter} & \multicolumn{2}{c}{\textbf{NSCLC}} & \multicolumn{2}{c}{\textbf{HNSCC}}\\
& CT 1 & CT 2 & CT 1 & CT 2\\
\midrule
number & 31 & 31 & 19 & 19\\
scanner & GE Healthcare Lightspeed 16 & GE Healthcare Lightspeed 16 & Siemens Biograph 16 & Siemens Biograph 16\\
 & GE Healthcare VCT & GE Healthcare VCT &  & \\
tube voltage (kVp) & 120 & 120 & 120 & 120\\
%tube current (mA) & 358 (200-441) & 360 (282-441) & 109 (156-187) & 29 (27-33)\\
exposure (mAs) & 8 (4-10) & 8 (4-13) & 36 (18-62) & 9 (9-10)\\
%exposure time (ms) & 478 (478-649) & 478 (478-649) & 500 & 500\\
reconstruction kernel & Lung & Lung & B31f & B19f\\
voxel spacing ($x$-axis; mm) & 0.67 (0.51-0.90) & 0.67 (0.51-0.91) & 0.98 & 1.37\\
voxel spacing ($y$-axis; mm) & 0.67 (0.51-0.90) & 0.67 (0.51-0.91) & 0.98 & 1.37\\
voxel spacing ($z$-axis; mm) & 1.25 & 1.25 & 3.00 (2.00-3.00) & 2.00\\
image noise ($\sigma$; HU) & 29.3 (16.6-76.1) & 28.4 (16.9-70.3) & 4.1 (3.9-5.4) & 4.2 (3.7-6.8)\\
% GTV volume (cm\textsuperscript{3}) & 13.2 (0.5-167.0) & 13.8 (0.4 - 167.0) & 70.4 (16.4-415.6) & 69.4 (16.1-413.2)\\
\bottomrule
\end{tabu}
\caption{Image acquisition parameters and characteristics for both NSCLC and HNSCC image data sets. Parameters were determined from the CT slices that contain portions of the gross tumour volume (GTV) region of interest. Numeric parameters are presented as \textit{median} (\textit{range}), unless only one value was found within the cohort. Image noise was calculated using Chang's method\cite{Chang2000} and represented by its standard deviation $\sigma$ (supplementary note 4). 
%Volume of the GTV was calculated by counting the number of voxels in the GTV mask and multiplying by the volume of each voxel.
\textit{kVp}: peak kilovoltage; \textit{HU}: Hounsfield unit}
\label{table:image_acquisition_parameters}
\end{table}

\newpage
\section*{Supplementary note 2: pre-interpolation low-pass filtering}
Image features are computed from voxels with uniform dimensions. In this work, features are computed with voxel spacings of $1$, $2$, $3$ and $4$ mm. The in-plane original spacing of the CT images is between $0.51$ and $1.37$ mm. We therefore need to down-sample images, which may cause image artefacts through aliasing and thus reduce feature robustness. In signal analysis, a signal may contain only frequencies up to half the sample frequency (the Nyquist frequency $\omega_N$) of the down-sampled signal to avoid artefacts. Signals are therefore low-pass filtered before down-sampling to suppress high frequency contents. The same concept applies to images as well. However, application of low-pass filters in radiomics is often neglected, despite the beneficial effect on feature robustness\cite{Mackin2018}.

We use a low-pass Gaussian filter before interpolation \texttt{scipy.ndimage.gaussian\_filter}. The Gaussian function $g(x)$ is defined as:
\begin{displaymath}
g(x)=\frac{1}{\sigma\sqrt{2\pi}}\text{e}^{-\frac{x^2}{2\sigma^2}},
\end{displaymath}
with $\sigma$ the standard deviation, or width, of the distribution. $\sigma$ is an input parameter for the Gaussian filter for which optimal settings have not been established. $\sigma$ moreover needs to be defined with respect to the typically non-uniformly spaced coordinate grid system of the original image and is thus specified separately for each axis.

Fourier theory allows us to set $\sigma$ based on the Nyquist frequency. The Fourier transform of the Gaussian function $g$ is\cite{Derpanis2005}:
\begin{displaymath}
G(\omega)=\text{e}^{-\frac{\omega^2 \sigma^2}{2}},
\end{displaymath}
with $\omega$ being a frequency. An ideal low-pass filter will maintain all frequencies $\omega < \omega_N$, and remove frequencies $\omega \geq \omega_N$ completely. However, ideal filters do not exist and a compromise is required between the desired attenuation of high-frequency content and the unwanted attenuation of low-frequency content. We define a smoothing parameter $\beta$, with $0<\beta\leq 1$, for the Fourier transformed Gaussian at $\omega=\omega_N$:
\begin{equation}\label{eq:nyquist_beta}
G(\omega_N)=\text{e}^{-\frac{\omega_N^2 \sigma^2}{2}}=\beta
\end{equation}

The Nyquist frequency $\omega_N$ may be expressed in terms of voxel spacing. For instance, we have a one-dimensional array of voxels with spacing $d_1$. We want to sample this array to spacing $d_2$. The sampling frequency is then $\omega_s=d_1 / d_2$, which leads to the Nyquist frequency $\omega_N=\omega_s/2=d_1/\left(2d_2\right)$.

We now solve equation (\ref{eq:nyquist_beta}) for $\sigma$:
\begin{gather*}
\text{e}^{-\frac{{\omega_N}^2 \sigma^2}{2}}=\beta \Leftrightarrow\\
\ln(\beta)=-\frac{{\omega_N}^2\sigma^2}{2} \Leftrightarrow\\
\sigma^2 = -\frac{2\ln(\beta)}{{\omega_N}^2} = -8\left(\frac{d_2}{d_1}\right)^2 \ln(\beta)  \Leftrightarrow \\
\sigma = -2\frac{d_2}{d_1}\sqrt{2\ln(\beta)}
\end{gather*}

We assess different parameter settings for $\beta$, namely $\beta=\lbrace0.50, 0.70, 0.80, 0.85, 0.90, \\0.93, 0.95, 0.97\rbrace$, as well as no low-pass filtering. Test-retest intraclass correlation coefficients (ICC (1,1)) and their 95\% confidence intervals (CI) are calculated on both test-retest cohorts \cite{Shrout1979}. The ICCs are used to determine the number of robust features and to show the ICC distribution. In addition, the distribution of the width of the ICC 95\% confidence intervals is assessed.

\begin{figure}[p]
\centering
\includegraphics[width=0.8\textwidth]{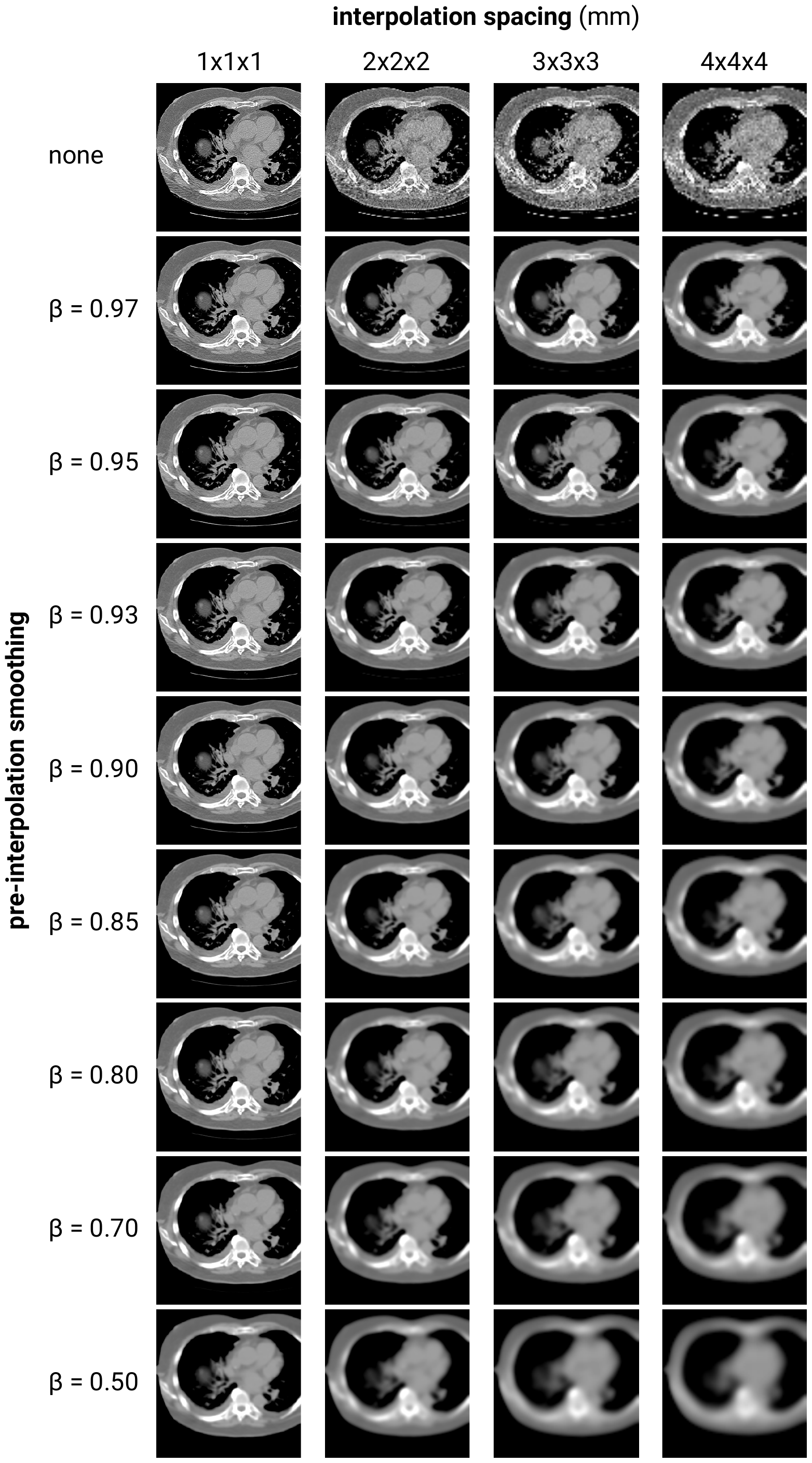}
\caption{Effect of smoothing and interpolation on a CT slice of an NSCLC patient. A Gaussian smoothing filter for the given $\beta$-values was applied before interpolation. Afterwards, tri-linear interpolation was conducted to resample to uniform voxel spacing (in mm). All slices are shown at the same size for comparison, and intensities were windowed between $[-400, 300]$ HU.}
\label{fig:beta_nsclc}
\end{figure}

\begin{figure}[p]
\centering
\includegraphics[width=0.8\textwidth]{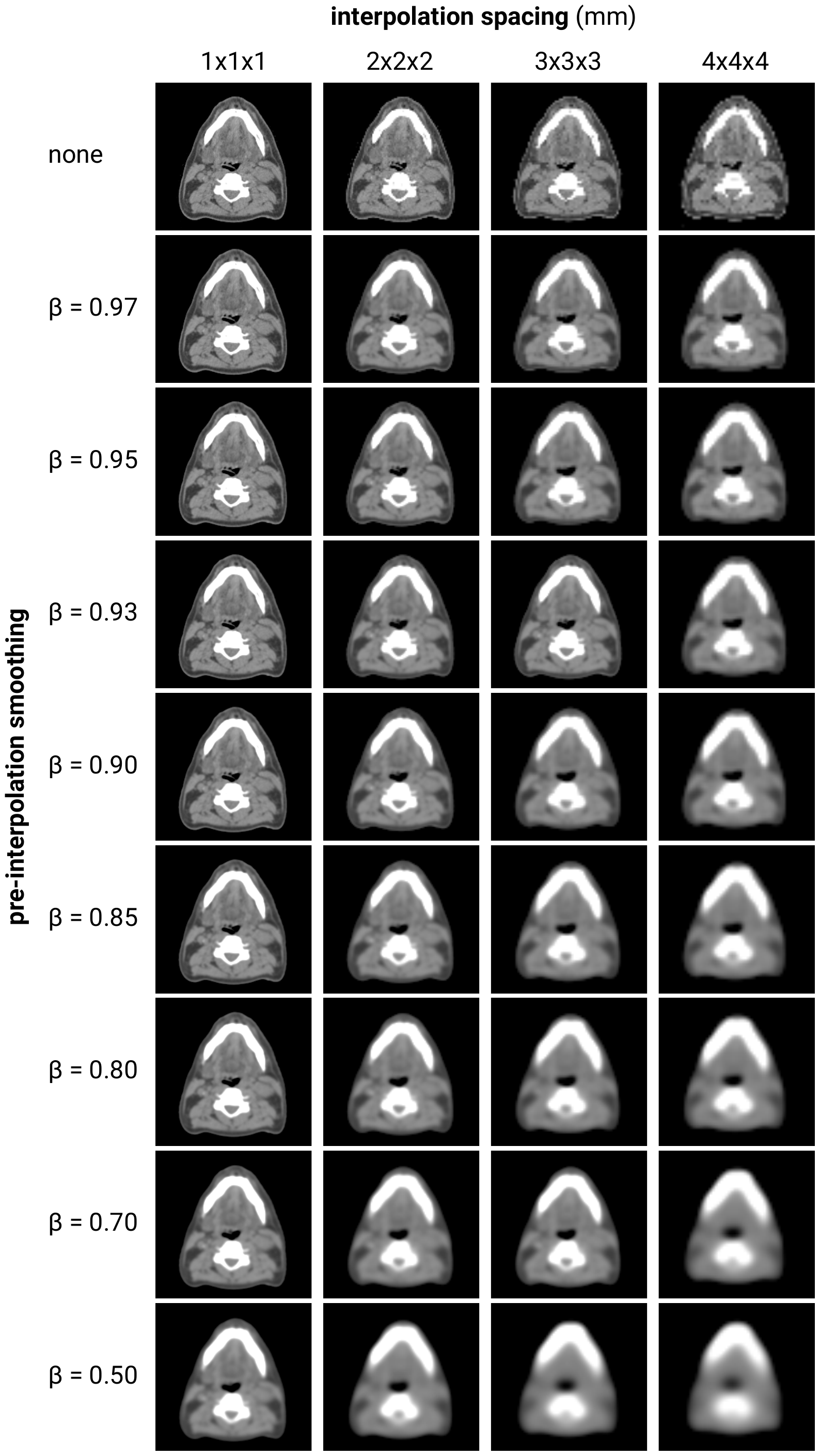}
\caption{Effect of smoothing and interpolation on a CT slice of an HNSCC patient. A Gaussian smoothing filter was applied before interpolation for the given $\beta$-values. Afterwards, tri-linear interpolation was conducted to resample to uniform voxel spacing (in mm). All slices are shown at the same size for comparison, and intensities were windowed between $[-220, 250]$ HU.}
\label{fig:beta_hnscc}
\end{figure}

Example images of an interpolated slice acquired from an NSCLC and an HNSCC patient are shown in Figures \ref{fig:beta_nsclc} and \ref{fig:beta_hnscc}, respectively. Down-sampling without interpolation caused visible image artefacts. On the other hand, images that are smoothed with a wide Gaussian low-pass filter (low $\beta$ value) lack detail.

\begin{figure}[ht]
\textbf{a}\\
\includegraphics[width=1.00\textwidth]{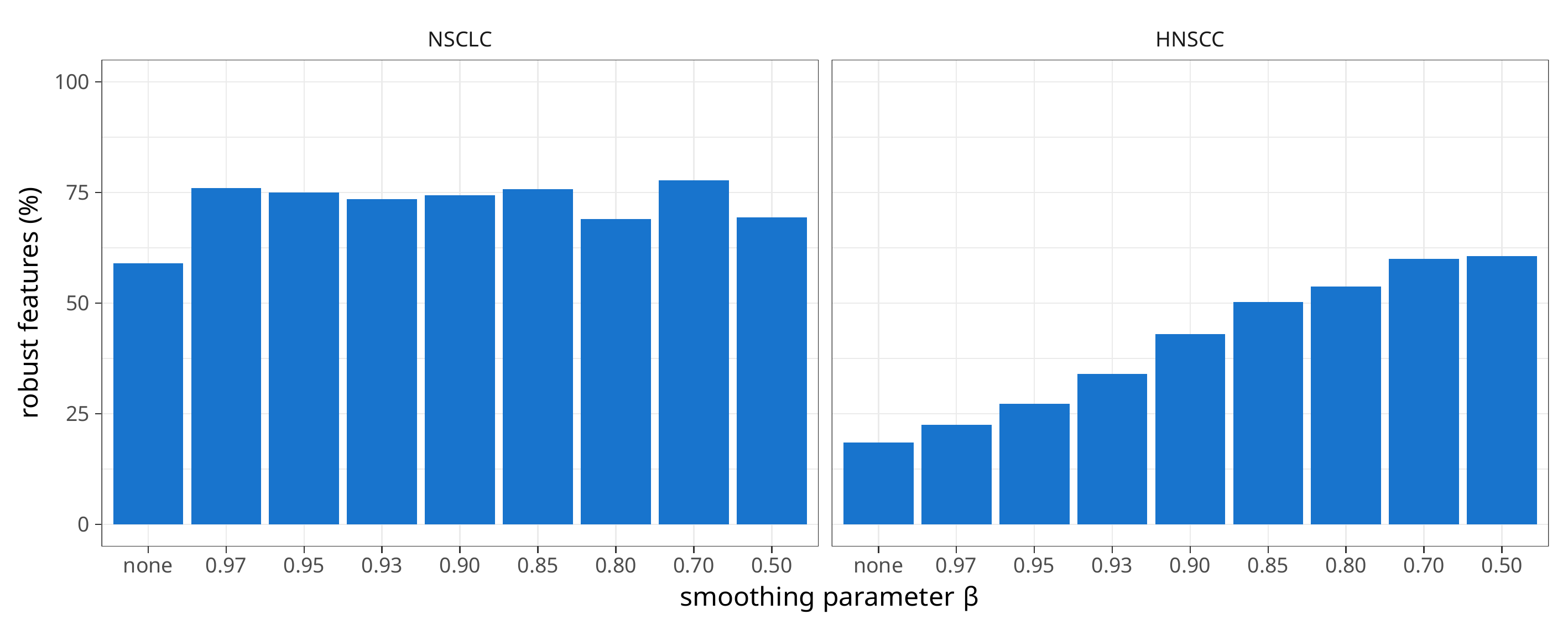}\\
\textbf{b}\\
\includegraphics[width=1.00\textwidth]{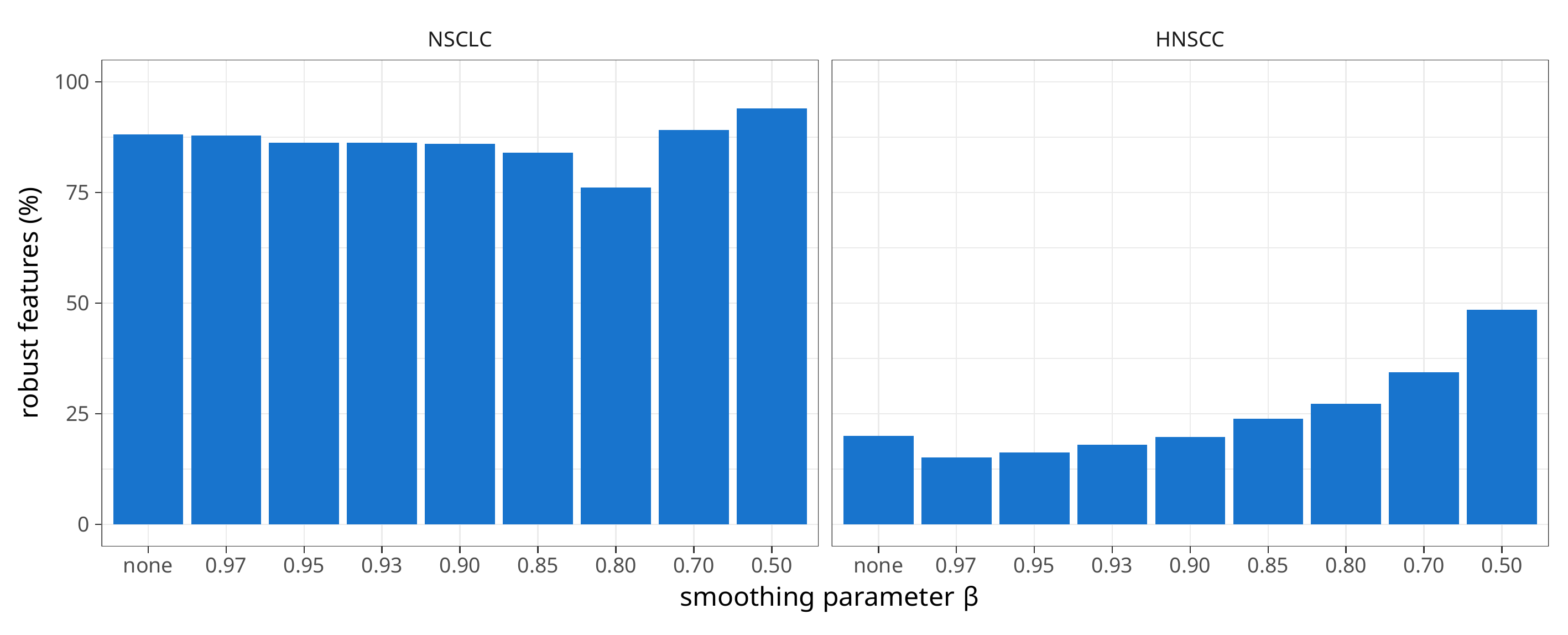}
\caption{Fraction of robust features according to the test-retest intraclass correlation coefficient (ICC (1,1)) for a pre-interpolation Gaussian smoothing parameter $\beta$. A feature was considered robust if ICC $\geq 0.90$. Lower $\beta$-values indicate stronger smoothing. The fraction of robust features is shown for all features (\textbf{a}) and for features acquired using a uniform spacing of 1 mm (\textbf{b}).}
\label{fig:beta_robust_features}
\end{figure}

\begin{figure}[ht]
\textbf{a}\\
\includegraphics[width=1.00\textwidth]{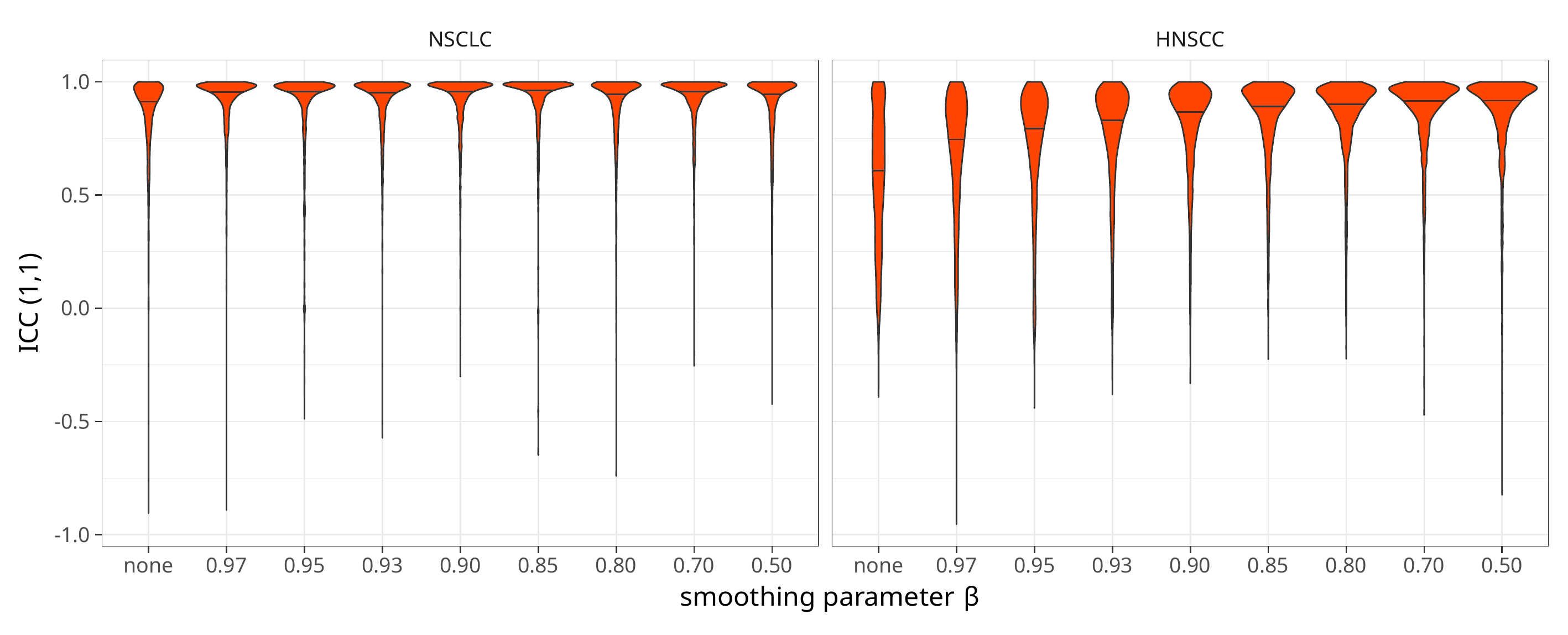}\\
\textbf{b}\\
\includegraphics[width=1.00\textwidth]{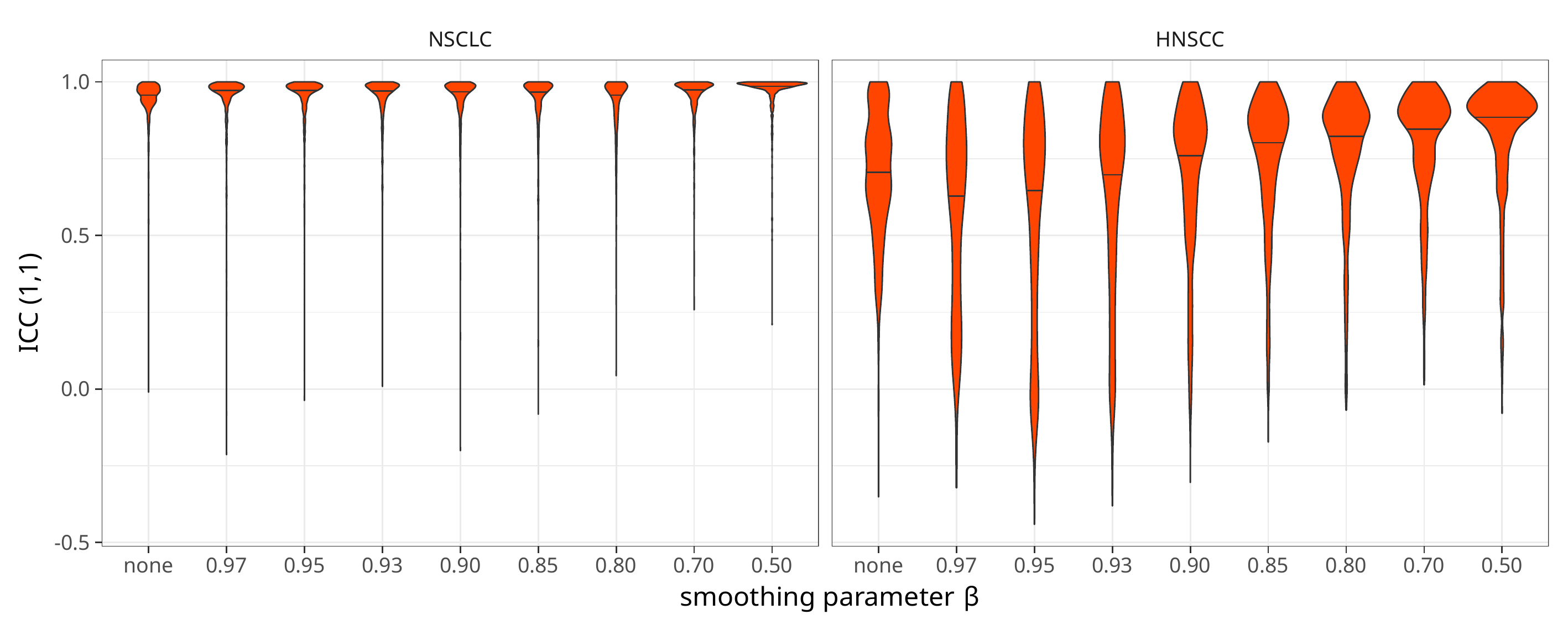}
\caption{Distribution of test-retest intraclass correlation coefficients (ICC (1,1)) for a pre-interpolation Gaussian smoothing parameter $\beta$. Lower $\beta$-values indicate stronger smoothing. The areas of the distributions were normalised. The median ICC in each distribution is indicated by a horizontal line. ICC distributions are shown for all features (\textbf{a}) and for features acquired using a uniform spacing of 1 mm (\textbf{b}).}
\label{fig:beta_icc}
\end{figure}

\begin{figure}[ht]
\textbf{a}\\
\includegraphics[width=1.00\textwidth]{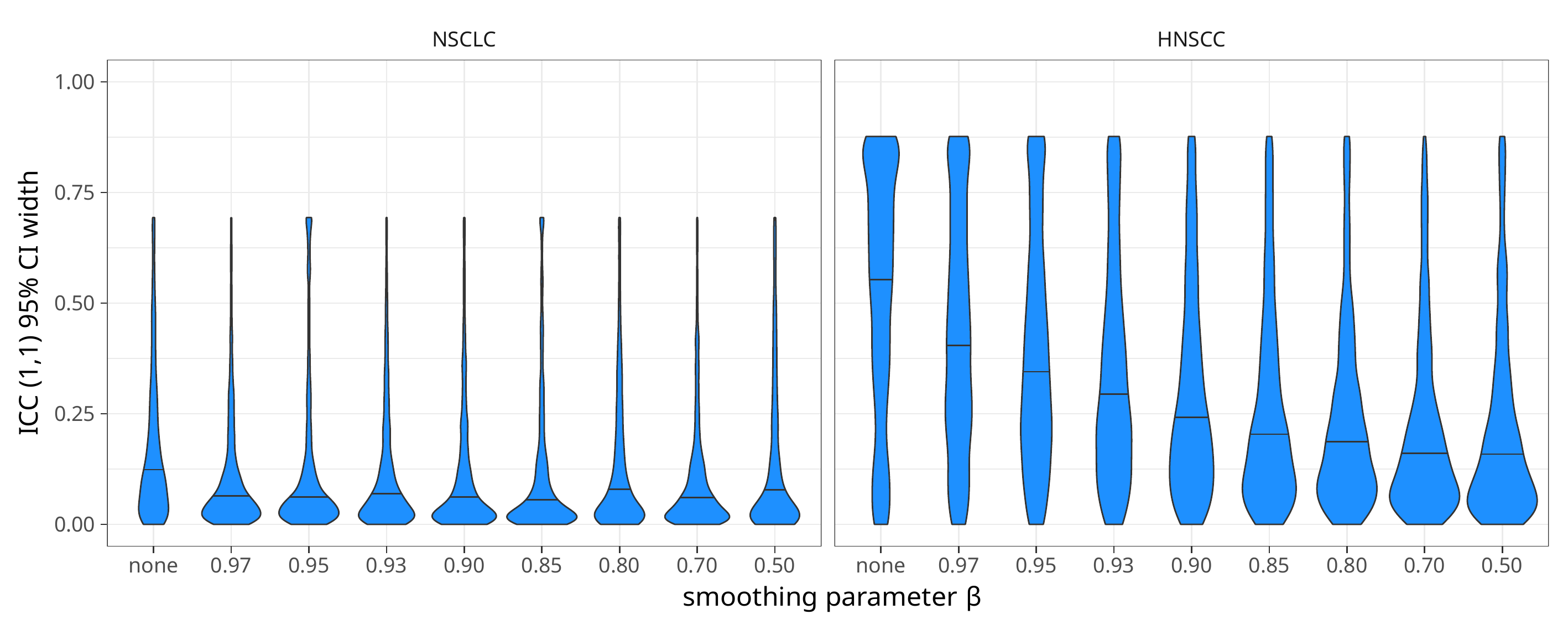}\\
\textbf{b}\\
\includegraphics[width=1.00\textwidth]{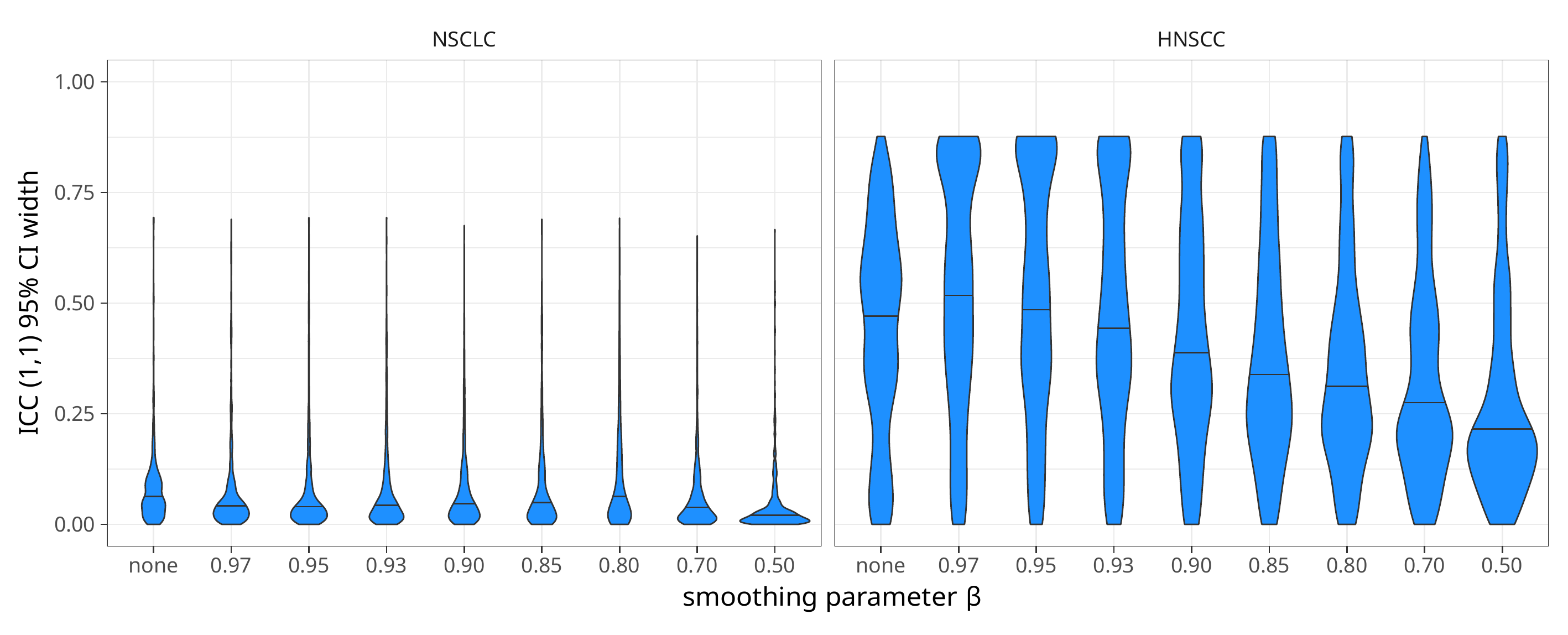}
\caption{Distribution of the 95 \% confidence interval (CI) widths of the test-retest intraclass correlation coefficients (ICC (1,1)) for a pre-interpolation Gaussian smoothing parameter $\beta$. Higher CI widths indicate larger variance in feature values between test and retest images. Lower $\beta$-values indicate stronger smoothing. The areas of the distributions were normalised. The median 95\% CI width is indicated in each distribution by a horizontal line. ICC CI width distributions are shown for all features (\textbf{a}) and for features acquired using a uniform spacing of 1 mm (\textbf{b}).}
\label{fig:beta_icc_width}
\end{figure}

The percentage of robust features according to the test-retest ICC is shown in Figure \ref{fig:beta_robust_features}. For the NSCLC cohort, even very light smoothing ($\beta = 0.97$) increases the percentage of robust features from 59.0\% to 75.9\%. With lower $\beta$-values, this percentage does not change, nor does the distribution of ICCs (Figure \ref{fig:beta_icc}) or the distribution of ICC CI widths (Figure \ref{fig:beta_icc_width}). For very low $\beta$-values, the ICC distribution for NSCLC may be less stable.

For the HNSCC cohort, the percentage of robust features increases with decreasing $\beta$, which is also reflected in the ICC distribution. In particular, even very mild smoothing ($\beta = 0.97$) increased the median ICC from 0.63 to 0.76. When only features computed with minimal down-sampling are considered (1 mm), $\beta=0.97$ reduced the median ICC from 0.72 to 0.65, and only recovered at $\beta=0.93$.  The same may be observed for the ICC CI width, which was increased for $\beta = 0.97$. A smoothing parameter value between $\beta = 0.93$ (robust features: 34.0\%; median ICC: 0.85; median CI width: 0.29) and $\beta = 0.90$ (robust features: 43.0\%; median ICC: 0.88; median CI width: 0.23) offers a good compromise between aliasing and lack of image details.

\FloatBarrier

\FloatBarrier

\newpage
\section*{Supplementary note 3: image features}
All image features were extracted according to the definitions provided by the Image Biomarker Standardisation Initiative\cite{Zwanenburg2016}. Intensity-volume histogram-based features were calculated for the 10\textsuperscript{th}, 25\textsuperscript{th}, 50\textsuperscript{th}, 75\textsuperscript{th} and 90\textsuperscript{th} intensity and volume fraction percentiles. Moran's I index and Geary's C measure were approximated by repeatedly selecting 100 voxels from the ROI at random and computing these metrics until the standard error of the mean decreased below 0.002. A total of 4032 features were computed, see Table \ref{table:feature_families}.

\begin{table}[t]
\centering
\begin{tabu}{@{}X[5,l,p] X[c,p] X[c,p] X[c,p]@{}}
\toprule
\textbf{feature family} & \textbf{count} & \textbf{multipl.} & \textbf{total}\\
\midrule
morphology & 29 & & 29 \\
local intensity & 2 & & 2 \\
intensity-based statistics & 18 &  & 18\\
intensity histogram & 23 & $\times 8$ & 184\\
intensity-volume histogram & 15 & & 15\\
grey level co-occurrence matrix & 25 & $\times 8$ & 200 \\
grey level run length matrix & 16 & $\times 8$ & 128\\
grey level size zone matrix & 16 & $\times 8$ &  128\\
grey level distance zone matrix & 16 & $\times 8$ & 128\\
neighbourhood grey tone difference matrix & 5 & $\times 8$ & 40\\
neighbouring grey level dependence matrix & 17 & $\times 8$ & 136\\
\midrule
\textit{total} & & & 1008 \\
\bottomrule
\end{tabu}
\caption{Feature families and the number of computed features. Several features require discretisation prior to computation. As two discretisation methods with each four bin size settings were evaluated, the number of such features is multiplied by 8. The final amount of features is 4032, due to calculation of each feature for four different interpolation spacings.}
\label{table:feature_families}
\end{table}

The following specific parameters were used to compute image features:
\begin{itemize}
\item \textbf{Morphology}: the surface mesh was constructed using the \textit{Marching Cubes} algorithm, with an iso-level of 0.5 \cite{Lorensen1987, Lewiner2003}.
\item \textbf{Intensity-volume histogram}: the intensity volume histogram was constructed as for images with discrete, defined (non-arbitrary) image values \cite{Zwanenburg2016}.
\item \textbf{Grey level co-occurrence matrix}: grey level co-occurrence matrices (GLCM) were calculated in 3D for 13 directions, with Chebyshev distance $\delta=1$. GLCM were symmetric and not distance-weighted. GLCM features were first calculated for every GLCM, and subsequently averaged.
\item \textbf{Grey level run length matrix}: grey level run length matrices (GLRLM) were calculated in 3D for 13 directions. GLRLM were not distance-weighted. GLRLM features were first calculated for every GLRLM, and subsequently averaged.
\item \textbf{Grey level size zone matrix}: a single grey level size zone matrix was calculated for the entire 3D volume.
\item \textbf{Grey level distance zone matrix}: a single grey level distance zone matrix was calculated for the entire 3D volume.
\item \textbf{Neighbourhood grey tone difference matrix}: a single neighbourhood grey tone difference matrix was calculated for the entire 3D volume, with Chebyshev distance $\delta=1$.
\item \textbf{Neighbouring grey level dependence matrix}: a single neighbouring grey level dependence matrix was calculated for the entire 3D volume, with Chebyshev distance $\delta=1$ and coarseness parameter $\alpha=0$.
\end{itemize}

Intensity histogram, grey level co-occurrence matrix, grey level run length matrix, grey level size zone matrix, grey level distance zone matrix, neighbourhood grey tone difference matrix and neighbouring grey level dependence matrix features required image discretisation prior to computation, which was conducted using two methods, with four settings each: a fixed bin number method with 8, 16, 32 or 64 bins; or a fixed bin width method with bins that were 6, 12, 18 or 24 Hounsfield units (HU) wide. For the fixed bin number method, the edge of the first bin coincided with the lowest intensity of the voxels included in the intensity mask. For the fixed bin width method, the lower edge of the first bin coincided with the lower edge of the re-segmentation range applied during image processing (NSCLC: -300 HU; HNSCC: -150 HU).

\FloatBarrier

\newpage
\section*{Supplementary note 4: Image perturbation algorithms}
This note provides additional information with regards to the implementation of the image perturbation algorithms. The algorithms were implemented in Python 3.6.1 (Python Software Foundation, Beaverton, Oregon, USA, \url{https://www.python.org/}). The implementation drew on functionality offered by the following libraries:
\begin{itemize}
\item \texttt{NumPy  1.13.3}\cite{Oliphant2007,Millman2011}, referred to as \texttt{numpy}.
\item \texttt{SciPy 0.19.1}\cite{Oliphant2007,Millman2011}, referred to as \texttt{scipy}.
\item \texttt{scikit-image 0.13.1}\cite{scikit-image}, referred to as \texttt{skimage}.
\item \texttt{PyWavelets 0.5.2}\cite{Lee2006}, referred to as \texttt{pywt}.
\end{itemize}
Specific functions from the libraries mentioned above are referred to in text.

\subsection*{Rotation}
Image rotations emulate changes due to different patient positioning. Image features should be robust against such perturbations to be reproducible.

The image is rotated in-plane around the $z$-axis by an angle $\theta$. Rotation was performed using the \texttt{scipy.ndimage.rotate} function, which implements rotation as an affine transformation. Bi-linear sampling is used to determine intensities in the rotated image. After rotation, intensities are rounded to the nearest integer value to conform with the expected integer Hounsfield units in CT.

The ROI mask is rotated in the same way as the image. However, the threshold for partial volume fractions in the mask is only applied after the interpolation step in the image processing scheme.

\subsection*{Noise}
Noise affects voxel intensities. Reproducible features should be robust to the noise present in an image. Perturbation by noise addition therefore follows two steps. First, the noise-dependent intensity variance is determined. Secondly, noise drawn from a Gaussian distribution with the same variance is added to the image.

The method of Chang \textit{et al.}\cite{Chang2000,Ikeda2010} is used to determine noise variance. In short, the image $\mathbf{I}$ is filtered in both the $x$ and $y$ direction in the image plane ($z$ being the axis along which the image slices are stacked) using a one-dimensional stationary \textit{coiflet-1} wavelet high-pass filter, \texttt{pywt.Wavelet("coif1").dec\_hi}. The filter convolution was implemented using the \texttt{scipy.ndimage.convolve1d} function. This cascade filter operation yields $\mathbf{I_{\text{diff}}}$. Subsequently, the noise level is estimated as:
\begin{displaymath}
\sigma_{\text{noise}}=\frac{\text{median}\left(\left| \mathbf{I_{\text{diff}}} \right| \right)}{0.6754}.
\end{displaymath}

Subsequently, for every image voxel random noise from a normal (Gaussian) distribution with mean $0$ and standard deviation $\sigma=\sigma_{\text{noise}}$ is generated (\texttt{numpy.random.normal}), and added. After noise addition, intensities are rounded to the nearest integer value to conform with the expected integer Hounsfield units in CT.

Noise variance is determined on the original image data, before any rotation, translation or other operation occurs. In the image processing scheme, noise addition takes place after rotation of the image, if applicable.

\subsection*{Translation}
Translation, like rotation, emulates changes due to different patient positioning. Translation was performed concurrently with interpolation, i.e. the interpolation grid was shifted off-centre by the provided translation fraction $\eta$ multiplied by the interpolation grid spacing. Translation was conducted along the $x$, $y$ and $z$ axes. Translation and interpolation was conducted with tri-linear approximation using the \texttt{scipy.ndimage.map\_coordinates} function.

\subsection*{Volume adaptation}
Shrinking or growing the segmentation mask is a method to mimic variance in expert delineations. For example, Fotina \textit{et al}. reported a mean coefficient of variance in volume of 14.9\% (range:$[4.4, 29.3]$\%) in CT-based expert delineations for lung and prostate cancer. The proposed method for volume adaptation is simple and intensity-agnostic, and is conducted as follows:
\begin{enumerate}
\item Approximate the volume $V_0$ of the ROI $\mathbf{R}_0$ by counting the number of voxels in the mask.
\item Calculate the volume of the ROI after adaptation (rounded down towards the nearest integer) by $V_a = \left\lfloor V_0 (1+\tau) \right\rfloor$, with $\tau$ the required growth/shrinkage fraction. $\tau>0.0$ indicates volume growth, and $\tau<0.0$ indicates shrinkage.
\item Define a geometric structure element that includes all voxels within Manhattan distance 1 (i.e. a centre voxel and its directly adjacent neighbours). We used the \texttt{scipy.ndimage.generate\_binary\_structure(3,1)} function.
\item Initialise a place-holder for an adapted mask $\mathbf{R}_p$ with volume $V_p$ by copying the original ROI and its volume. This place-holder is used to track the volume and mask over iterative adaptations.
\item Iterate the mask shrinkage/growth process until the loop breaks:
\begin{enumerate}
\item If $\tau > 0.0$ dilate the mask (\texttt{scipy.ndimage.binary\_dilation}) once, using the structure element defined in step 3.
\item If $\tau < 0.0$ erode the mask (\texttt{scipy.ndimage.binary\_erosion}) once, using the structure element defined in step 3.
\item Approximate the volume $V_n$ of the newly adapted mask $\mathbf{R}_n$ by counting the number of voxels in the mask.
\item If $V_n=0.0$ break from the loop.
\item If $\tau > 0.0$ and $V_n > V_a$ break from the loop.
\item If $\tau < 0.0$ and $V_n < V_a$ break from the loop.
\item Replace the previous place-holder mask by setting $\mathbf{R}_p=\mathbf{R}_n$. This is done until the final growth/shrinkage iteration, when one of the conditions in steps d-f was satisfied. 
\end{enumerate}
\item If $V_n\neq V_a$, $\mathbf{R}_n$ contains either too many ($\tau > 0.0$) or too few ($\tau < 0.0$) voxels. A limited number of voxels should be added to or removed from the mask $\mathbf{R}_p$ to complete the adaptation. Practically, we update the rim formed by the disjunctive union of $\mathbf{R}_p$ and $\mathbf{R}_n$, i.e. $\mathbf{R}_r=\mathbf{R}_n \ominus \mathbf{R}_p$:
\begin{enumerate}
\item Determine the number of voxels to be added/removed from the mask: $N = |V_a - V_p|$.
\item Find rim $\mathbf{R}_r$ by logical \texttt{XOR} comparison of $\mathbf{R}_n$ and $\mathbf{R}_p$ (\texttt{numpy.logical\_xor}).
\item Select $N$ voxels from the rim at random, without replacement (\texttt{numpy.random.choice}).
\item If $N>0$ and $\tau>0.0$ add the $N$ voxels to mask $\mathbf{R}_p$.
\item If $N>0$ and $\tau<0.0$ remove the $N$ voxels from mask $\mathbf{R}_p$.
\end{enumerate}
\item Volume adaptation ends. The mask $\mathbf{R}_p$ defines the perturbed region of interest.
\end{enumerate}

\subsection*{Contour randomisation}
Multiple image segmentations are required for randomising the contour of the region of interest. Creating multiple segmentations usually requires delineation by multiple experts. However, for larger quantities of image data, the creation of multiple manual delineations is extremely time-consuming and unfeasible in practice. An automated contour randomisation is therefore required. We use supervoxel-based segmentation algorithm for randomising contours. Supervoxels are connected clusters of voxels with similar intensity characteristics. To create a random contour, we compare supervoxels with a single segmentation delineated by an expert. The region of interest (ROI) is then randomised based on the overlap of supervoxels with the expert contour. Multiple algorithms produce supervoxels. We used the \textit{simple linear iterative clustering} (SLIC) algorithm as it efficiently produces compact, contiguous supervoxels \cite{Achanta2012}. This algorithm was provided through the \texttt{skimage.segmentation.slic\_superpixels} function.

\noindent Contour randomisation is conducted as follows:
\begin{enumerate}
\item Both the image and the region of interest (ROI) mask are cropped to 25 mm around the ROI bounding box to limit computational costs.
\item The intensities of the cropped image stack $\mathbf{I}$ are translated to a $[0,1]$ range:
\begin{enumerate}
\item Intensities $I_j \in \mathbf{I} $ are first restricted to range $\mathbf{r}$, which is based on the range used for ROI re-segmentation (Table 2 in main manuscript). The intensity range extends the re-segmentation range by 10\% at both the upper ($g_u$) and lower ($g_l$) boundaries:
\begin{displaymath}
\mathbf{r}=\left[g_l - 0.1\cdot \left(g_u - g_l\right), g_u + 0.1\cdot \left(g_u - g_l\right)\right]=\left[r_1, r_2\right]
\end{displaymath}
All intensity values outside range $\mathbf{r}$ are replaced by the nearest valid intensity:
\begin{displaymath}
I_j =
\begin{cases}
r_1, & I_j < r_1 \\
r_2, & I_j > r_2 \\
I_j, & \text{otherwise}
\end{cases}
\end{displaymath}
\item Intensities are then mapped to the $[0,1]$ range by a simple transformation:
\begin{displaymath}
I_{j,s} = \frac{I_j - r_1}{r_2 - r_1}
\end{displaymath}
\end{enumerate}
\item The number of supervoxels is estimated so that on average each supervoxel occupies 0.5 cm\textsuperscript{3}:
\begin{displaymath}
N_{\text{sx}, \text{est}} = \left\lceil \frac{N_v\, V_{vox}} {0.5}\right\rceil,
\end{displaymath}
with $N_v$ the number of voxels in $\mathbf{I}_s$ and $V_{\text{vox}}$ the volume of each voxel (in cm\textsuperscript{3}). $\left\lceil\ldots\right\rceil$ denotes a ceiling operation that rounds the fraction up towards the nearest integer.
\item The SLIC algorithm pre-processes $\mathbf{I}_s$ by applying a Gaussian smoothing filter. The filter scaling parameter $\sigma$ is set to the uniform voxel spacing (1,2,3 or 4 mm).
\item SLIC is performed, using the \texttt{skimage.segmentation.slic\_superpixels} function, with filter scaling parameter $\sigma$, the estimated number of supervoxels $N_{\text{sx}, \text{est}}$, compactness $\beta=0.05$, and by allowing supervoxels to vary in size between 0.25 cm\textsuperscript{3} and 1.5 cm\textsuperscript{3}. This results in a mask $\mathbf{S}$ that labels supervoxels in $\mathbf{I}_s$. A total $N_{\text{sx}}$ supervoxels are labelled.
\item The overlap $\eta_k$ of the different supervoxels $\mathbf{S}_k \subset \mathbf{S}$, where $k=1,\ldots, N_{\text{sx}}$, and the morphological ROI mask $\mathbf{R}$ defined by the expert is determined as follows:
\begin{enumerate}
\item The number of voxels $m_k$ labelled by supervoxel $\mathbf{S}_k$ is counted.
\item The number of voxels $m_{\eta, k}$ in the intersection of the ROI mask and the supervoxel, $\mathbf{R} \cap \mathbf{S}_k$ is counted.
\item The overlap fraction for supervoxel $k$ is then defined as:
\begin{displaymath}
\eta_k = m_k / m_{\eta, k}
\end{displaymath}
By definition, $0.0 \leq \eta_k \leq 1.0$.
\end{enumerate}
\item Subsequently, supervoxels are selected to form a new supervoxel-based ROI mask $\mathbf{R}_{\text{sx}}$, as follows:
\begin{enumerate}
\item To ensure that the new ROI mask will not remain empty, i.e. $\mathbf{R}_{\text{sx}}\neq \emptyset$, the supervoxel with the highest overlap is always selected, regardless of the actual overlap.
\item Additionally, all supervoxels with overlap $\eta \geq 0.90$ are always selected.
\item All supervoxels with overlap $\eta < 0.20$ are never selected.
\item All supervoxels with overlap $0.20\leq \eta \leq 0.90$ are randomly selected. For each supervoxel $k$, a random number is drawn uniformly from the interval $[0,1]$, i.e. $x_k \sim \mathcal{U}(0,1)$, using the \texttt{numpy.random.random} function. If $x_k \leq \eta_k$, the supervoxel is added to the mask. Thus, selection probabilities for such supervoxels are equal to the overlap.
\item The resulting supervoxel-based ROI mask $\mathbf{R}_{\text{sx}}$ is morphologically closed using the \texttt{scipy.ndimage.binary\_closing} function with a geometric structure element that includes all voxels within Manhattan distance 1 (i.e. a centre voxel and its directly adjacent neighbours).
\end{enumerate}
\item Contour randomisation ends. The mask $\mathbf{R}_{\text{sx}}$ defines the perturbed region of interest.
\end{enumerate}

\newpage
\section*{Supplementary note 5: image perturbation settings}

Eighteen perturbation chains were constructed from the five basic perturbations. Rotation was performed by rotating the image around the $z$-axis by an angle $\theta$. Translation was performed by shifting the voxel grid by a fraction $\eta$ of the voxel spacing. If more than one fraction was provided, translations were performed using all permutations of $\eta$ and the three primary axes. Thus, we performed eight permutations if two fractions $\eta$ were provided, and 27 for three fractions. Volume adaptation required a shrinkage/growth fraction $\tau$, with negative values indicating shrinkage and positive values growth of the mask. Noise adaptation and contour randomisation did not require additional settings, but could be repeated.

We perturbed images using every permutation of the settings of a perturbation chain. Several combinations of perturbations were not tested as the number of permutations was excessive. In particular, chains that combined rotation, translation and volume adaptation were not tested, as a typical set of 5 rotation angles, 2 translation fractions and 5 volume growth/shrinkage factors would lead to 200 permutations. We defined perturbation chains that lead to roughly 30 permutations to limit the effect of sample size on the intraclass correlation coefficient. In addition, we did not test every possible perturbation chain that included noise addition, as noise addition had a marginal effect if used in combination with other perturbations.

The following perturbations were defined, with $m$ the total number of perturbed images generated:
{\small
\begin{enumerate}
\item Rotation (R, $m=27$)
\begin{itemize}
\item \textit{rotation}: $\theta=\left\lbrace -13^{\circ}, -12^{\circ},\ldots ,13^{\circ}\right\rbrace$
\end{itemize}

\item Noise addition (N, $m=30$)
\begin{itemize}
\item \textit{noise addition}: 30 repetitions
\end{itemize}

\item Translation (T, $m=27$)
\begin{itemize}
\item \textit{translation}: $\eta=\left\lbrace 0.0, 0.333, 0.667\right\rbrace$
\end{itemize}

\item Volume adaptation (V, $m=29$)
\begin{itemize}
\item \textit{volume adaptation}: $\tau=\left\lbrace -0.28, -0.26,\ldots ,0.28\right\rbrace$
\end{itemize}

\item Contour randomisation (C, $m=30$)
\begin{itemize}
\item \textit{contour randomisation}: 30 repetitions
\end{itemize}

\item Rotation and translation (RT, $m=32$)
\begin{itemize}
\item \textit{rotation}: $\theta=\left\lbrace -6^{\circ}, -2^{\circ}, 2^{\circ}, 6^{\circ} \right\rbrace$
\item \textit{translation}: $\eta=\left\lbrace 0.25, 0.75\right\rbrace$
\end{itemize}

\item Rotation, noise addition and translation (RNT, $m=32$)
\begin{itemize}
\item \textit{rotation}: $\theta=\left\lbrace -6^{\circ}, -2^{\circ}, 2^{\circ}, 6^{\circ} \right\rbrace$
\item \textit{noise addition}: 1 repetition
\item \textit{translation}: $\eta=\left\lbrace 0.25, 0.75\right\rbrace$
\end{itemize}

\item Rotation and volume adaptation (RV, $m=30$)
\begin{itemize}
\item \textit{rotation}: $\theta=\left\lbrace -10^{\circ}, -6^{\circ}, -2^{\circ}, 2^{\circ}, 6^{\circ}, 10^{\circ}\right\rbrace$
\item \textit{volume adaptation}: $\tau=\left\lbrace-0.2, -0.1, 0.0, 0.1, 0.2\right\rbrace$
\end{itemize}

\item Rotation and contour randomisation (RC, $m=27$)
\begin{itemize}
\item \textit{rotation}: $\theta=\left\lbrace -13^{\circ}, -12^{\circ},\ldots ,13^{\circ}\right\rbrace$
\item \textit{contour randomisation}: 1 repetition
\end{itemize}

\item Translation and volume adaptation (TV, $m=40$)
\begin{itemize}
\item \textit{translation}: $\eta=\left\lbrace 0.25, 0.75\right\rbrace$
\item \textit{volume adaptation}: $\tau=\left\lbrace-0.2, -0.1, 0.0, 0.1, 0.2\right\rbrace$
\end{itemize}

\item Translation and contour randomisation (TC, $m=27$)
\begin{itemize}
\item \textit{translation}: $\eta=\left\lbrace 0.0, 0.333, 0.667\right\rbrace$
\item \textit{contour randomisation}: 1 repetition
\end{itemize}

\item Rotation, translation, and contour randomisation (RTC, $m=32$)
\begin{itemize}
\item \textit{rotation}: $\theta=\left\lbrace -6^{\circ}, -2^{\circ}, 2^{\circ}, 6^{\circ} \right\rbrace$
\item \textit{translation}: $\eta=\left\lbrace 0.25, 0.75\right\rbrace$
\item \textit{contour randomisation}: 1 repetition
\end{itemize}

\item Rotation, noise addition, translation, and contour randomisation (RNTC, $m=32$)
\begin{itemize}
\item \textit{rotation}: $\theta=\left\lbrace -6^{\circ}, -2^{\circ}, 2^{\circ}, 6^{\circ} \right\rbrace$
\item \textit{noise addition}: 1 repetition
\item \textit{translation}: $\eta=\left\lbrace 0.25, 0.75\right\rbrace$
\item \textit{contour randomisation}: 1 repetition
\end{itemize}

\item Volume adaptation and contour randomisation (VC, $m=30$)
\begin{itemize}
\item \textit{volume adaptation}: $\tau=\left\lbrace-0.2, -0.1, 0.0, 0.1, 0.2\right\rbrace$
\item \textit{contour randomisation}: 6 repetitions
\end{itemize}

\item Rotation, volume adaptation and contour randomisation (RVC, $m=30$)
\begin{itemize}
\item \textit{rotation}: $\theta=\left\lbrace -10^{\circ}, -6^{\circ}, -2^{\circ}, 2^{\circ}, 6^{\circ}, 10^{\circ}\right\rbrace$
\item \textit{volume adaptation}: $\tau=\left\lbrace-0.2, -0.1, 0.0, 0.1, 0.2\right\rbrace$
\item \textit{contour randomisation}: 1 repetition
\end{itemize}

\item Rotation, noise addition, volume adaptation and contour randomisation (RNVC, $m=30$)
\begin{itemize}
\item \textit{rotation}: $\theta=\left\lbrace -10^{\circ}, -6^{\circ}, -2^{\circ}, 2^{\circ}, 6^{\circ}, 10^{\circ}\right\rbrace$
\item \textit{noise addition}: 1 repetition
\item \textit{volume adaptation}: $\tau=\left\lbrace-0.2, -0.1, 0.0, 0.1, 0.2\right\rbrace$
\item \textit{contour randomisation}: 1 repetition
\end{itemize}

\item Translation, volume adaptation and contour randomisation (TVC, $m=40$)
\begin{itemize}
\item \textit{translation}: $\eta=\left\lbrace 0.25, 0.75\right\rbrace$
\item \textit{volume adaptation}: $\tau=\left\lbrace-0.2, -0.1, 0.0, 0.1, 0.2\right\rbrace$
\item \textit{contour randomisation}: 1 repetition
\end{itemize}

\item Noise addition, translation, volume adaptation and contour randomisation (NTVC, $m=40$)
\begin{itemize}
\item \textit{noise addition}: 1 repetition
\item \textit{translation}: $\eta=\left\lbrace 0.25, 0.75\right\rbrace$
\item \textit{volume adaptation}: $\tau=\left\lbrace-0.2, -0.1, 0.0, 0.1, 0.2\right\rbrace$
\item \textit{contour randomisation}: 1 repetition
\end{itemize}

\end{enumerate}
}

\newpage
\section*{Supplementary note 6: Robustness differences between perturbed test and retest images}
Perturbation ICCs are averaged between test and retest images for easier comparison with test-retest ICCs. To verify that there is no significant bias in perturbation ICC towards one image, we first calculated the difference between the perturbation ICCs of the same feature for every feature. Subsequently, we calculate the mean $\mu$ and standard deviation $\sigma$ of the differences, and perform a one-sided location test against mean 0:
\begin{displaymath}
z=\sqrt{n}\frac{\mu-0}{\sigma}
\end{displaymath}

\begin{figure}[ht]
\includegraphics[width=1.0\textwidth]{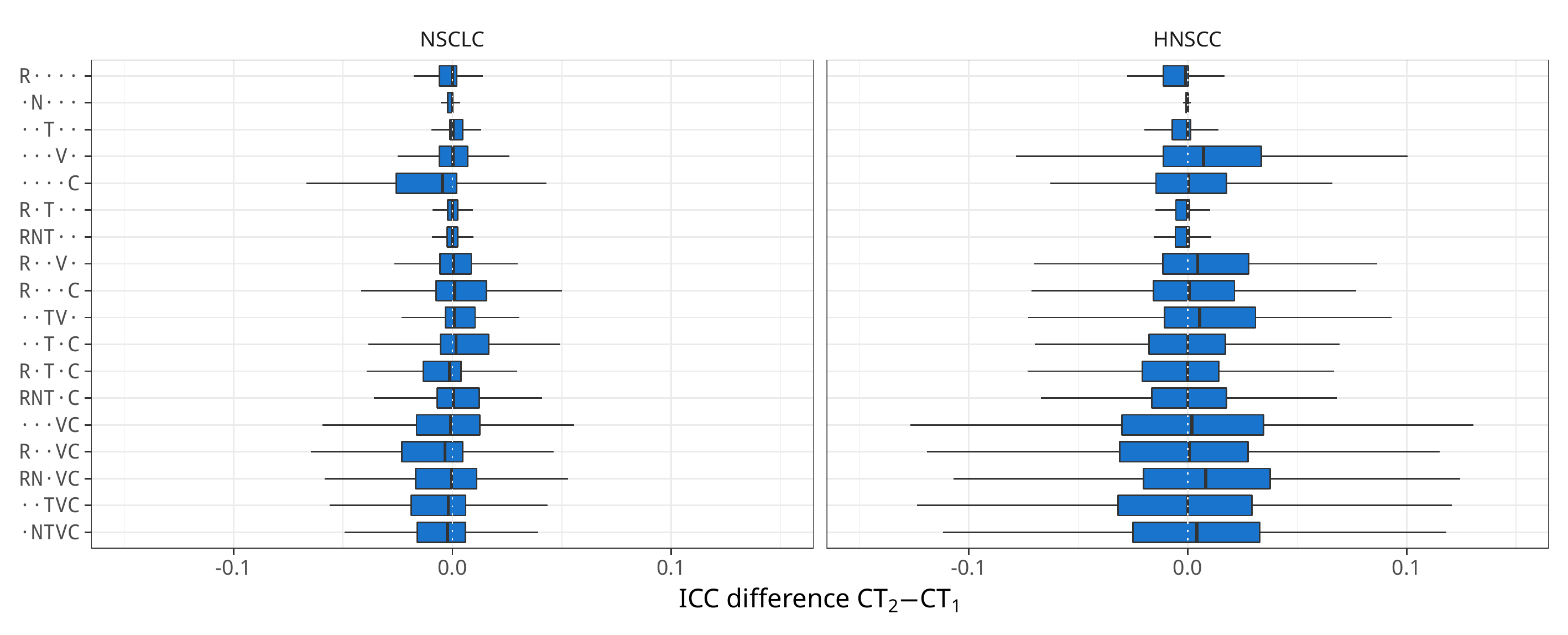}
\caption{Box plots of the differences in intraclass correlation coefficient (ICC) between test (CT\textsubscript{1}) and retest (CT\textsubscript{2}) data sets for the perturbation chains. The boxes cover the interquartile range (IQR), and the median ICC is indicated. The whiskers of each plot extend to 1.5 times the IQR.}
\label{fig:icc_differences}
\end{figure}

$|z|\geq 1.96$ corresponds to a significance level $p\leq 0.05$. The ICC difference of each feature can not be considered independent as many features are known to be correlated, which affects the choice for $n$. Hence, we chose $n=1$ (complete pooling), instead of $n=4032$ for independent samples ($Z$-test). None of perturbations were distributed significantly from 0. The distribution of perturbation ICC differences is shown in Figure \ref{fig:icc_differences}.

\newpage
\bibliography{reference}

\end{document}